\documentclass{article} 
\usepackage{iclr2025_conference,times}

\usepackage{hyperref}
\usepackage{url}
\usepackage{graphicx}
\usepackage{algorithm}
\usepackage{algorithmic}
\usepackage{booktabs}
\usepackage{enumitem}
\usepackage{wrapfig}
\usepackage{colortbl}
\usepackage{amsmath,amsfonts,bm}

\hypersetup{
    colorlinks,
    citecolor=[rgb]{0.00, 0.45, 0.70},
    linkcolor=[rgb]{0.01, 0.62, 0.45},
    urlcolor=[rgb]{0.80, 0.47, 0.74},
}

\definecolor{gray}{RGB}{225,225,225}
\definecolor{lred}{RGB}{255,235,235}

\title{Ensembling Diffusion Models via Adaptive Feature Aggregation}

\author{Cong Wang$^1$\thanks{Work done during an internship at Tencent AIPD.}~~\thanks{Equal contribution.}~~~~~~~~
Kuan Tian$^2$\footnotemark[2]~~~~~~~~
Yonghang Guan$^{2}$~~~~~~
Fei Shen$^2$~~~~~~
Zhiwei Jiang$^1$\thanks{Corresponding authors.}\\
\textbf{Qing Gu$^1$~~~~~~~~~~~~~~~~Jun Zhang$^2$\footnotemark[3]} \\[5pt]
$^1$ State Key Laboratory for Novel Software Technology, Nanjing University \\
$^2$ Tencent AIPD \\[3pt]
\texttt{cw@smail.nju.edu.cn, \{kuantian, yohnguan, ffeishen\}@tencent.com,} \\
\texttt{\{jzw, guq\}@nju.edu.cn, junejzhang@tencent.com}
}

\allowdisplaybreaks

\iclrfinalcopy 
\begin{document}

\maketitle
\begin{abstract}
The success of the text-guided diffusion model has inspired the development and release of numerous powerful diffusion models within the open-source community.
These models are typically fine-tuned on various expert datasets, showcasing diverse denoising capabilities. 
Leveraging multiple high-quality models to produce stronger generation ability is valuable, but has not been extensively studied.
Existing methods primarily adopt parameter merging strategies to produce a new static model. 
However, they overlook the fact that the divergent denoising capabilities of the models may dynamically change across different states, such as when experiencing different prompts, initial noises, denoising steps, and spatial locations. 
In this paper, we propose a novel ensembling method, Adaptive Feature Aggregation (AFA), which dynamically adjusts the contributions of multiple models at the feature level according to various states (i.e., prompts, initial noises, denoising steps, and spatial locations), thereby keeping the advantages of multiple diffusion models, while suppressing their disadvantages.
Specifically, we design a lightweight Spatial-Aware Block-Wise (SABW) feature aggregator that adaptive aggregates the block-wise intermediate features from multiple U-Net denoisers into a unified one.
The core idea lies in dynamically producing an individual attention map for each model's features by comprehensively considering various states.
It is worth noting that only SABW is trainable with about 50 million parameters, while other models are frozen. 
Both the quantitative and qualitative experiments demonstrate the effectiveness of our proposed method.\footnote{The code is available at \url{https://github.com/tenvence/afa}.}
\end{abstract}

\section{Introduction}

Diffusion models \citep{sohl2015deep,ho2020denoising} have progressively become the mainstream models for text-guided image generation \citep{nichol2021glide, ramesh2022hierarchical, saharia2022photorealistic, rombach2022high, balaji2022ediffi, xue2023raphael, feng2023ernie}, which treats generation as an iterative denoising task.
Recently, the open-source stable diffusion (SD) \citep{rombach2022high} model has prompted the development and release of numerous powerful diffusion models within the open-source community (e.g., CivitAI\footnote{\url{https://civitai.com/}}).
These models are typically fine-tuned on various expert datasets, showcasing diverse denoising capabilities.
Leveraging multiple high-quality models to dig out better generations is an important research direction, which has not been extensively studied.

\begin{figure}[t]
    \centering
    \includegraphics[width=\linewidth]{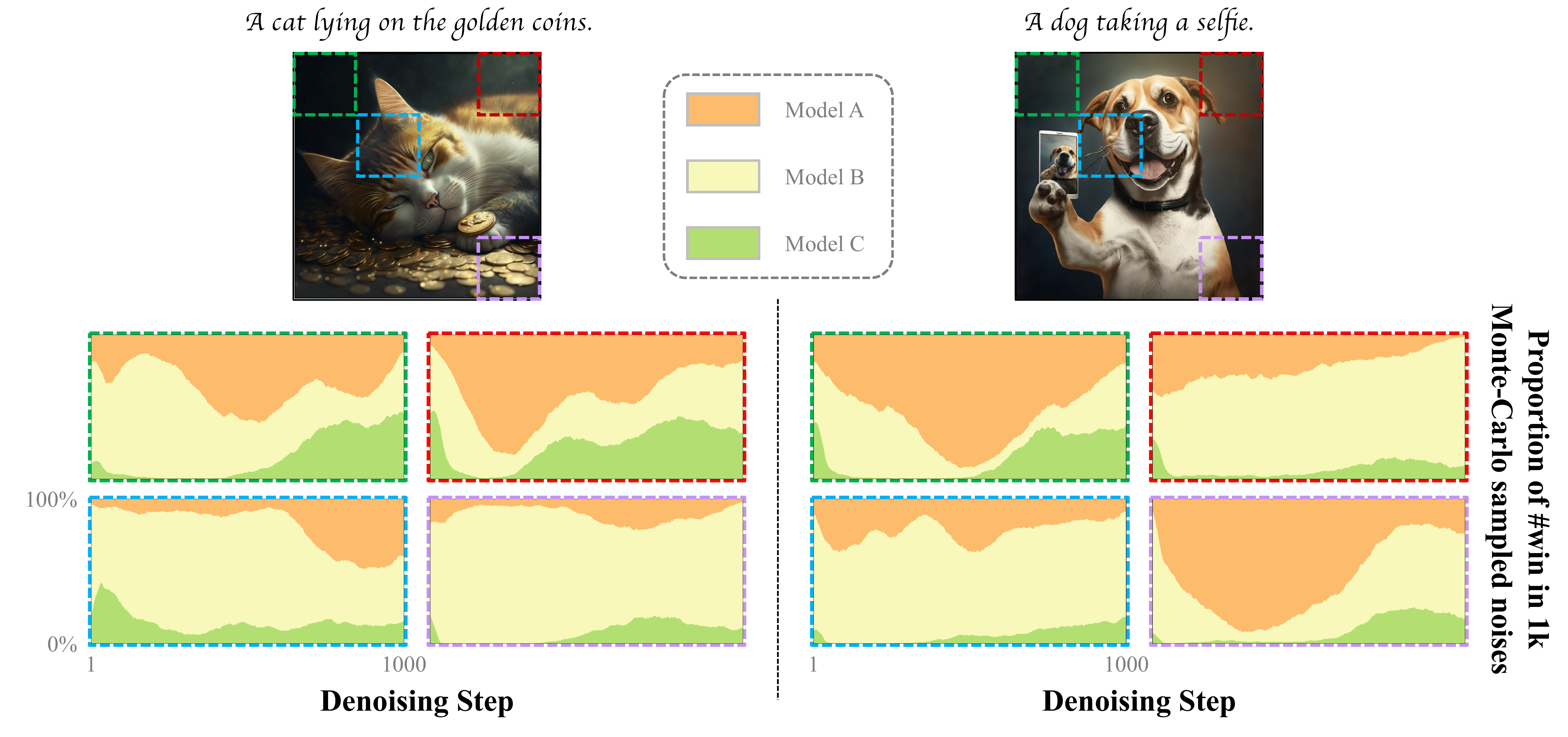}
    \vspace{-5pt}
    \caption{Examples to illustrate the dynamical change of the denoising capabilities across various states. We conduct experiments on different prompts with various initial noises. We then plot the proportion of \textit{wins} (i.e., the model with the least error between the predicted noise and the initial noise), for each model in a certain spatial region.}
    \vspace{-5pt}
    \label{fig:positional-denoising-capability}
\end{figure}

Existing methods leverage multiple diffusion models through the weighted merging of model parameters, which can be called the static method.
The weights are usually manually set (e.g., Weighted-Merging\footnote{\url{https://github.com/hako-mikan/sd-webui-supermerger}} and MBW\footnote{\url{https://github.com/bbc-mc/sdweb-merge-block-weighted-gui}}) or automatically searched through enumeration (e.g., autoMBW\footnote{\url{https://github.com/Xerxemi/sdweb-auto-MBW}}).
In contrast, model ensembling, which can be called the dynamic method, often uses dynamic strategies to fuse multiple models at the feature level. Unlike ensembling for classification models \citep{freund1995desicion} that usually work after the output, ensembling for diffusion models typically needs to work for each block. 
However, the denoising capabilities of models vary not only at different blocks but also at different spatial locations. 
As illustrated in Figure~\ref{fig:positional-denoising-capability}, we conduct denoising experiments on different prompts with various initial noises. Then, we plot the proportion of \textit{wins} (i.e., the model with the least error between the predicted noise and the initial noise), for each model in a certain spatial region, to distinguish the denoising capabilities of the models at different states. 
In other words, different prompts, initial noises, denoising steps, and spatial locations can all have a significant impact on the denoising capabilities of diffusion models.
This implies that an adaptive method is needed to ensure that each diffusion model dominates the generation at its strongest states.

In this paper, we propose a novel Adaptive Feature Aggregation (AFA) method, which dynamically adjusts the contributions of multiple models at the feature level by taking into account various states, such as prompts, initial noises, denoising steps, and spatial locations.
Specifically, we design a lightweight Spatial-Aware Block-Wise (SABW) feature aggregator that adaptively aggregates the block-level intermediate features from multiple U-Net denoisers into a unified one.
The core idea of adaptive aggregation lies in dynamically producing an individual attention map for each model's features by comprehensively taking into account the various states.
A noteworthy aspect of AFA is that only SABW is trainable with about 50 million parameters, while all other models are frozen.

Our main contributions are summarized as follows:
\begin{itemize}[leftmargin=1em]
    \item We propose an ensembling-based AFA method to dynamically adjust the contributions of multiple models at the feature level.
    \item We design the SABW feature aggregator that can produce attention maps according to various states to adaptively aggregate the block-level intermediate features from multiple U-Net denoisers.
    \item We conduct both quantitative and qualitative experiments, demonstrating that our AFA outperforms the base models and the baseline methods in both superior quality and context alignment.
\end{itemize}
 
\section{Related Work}

\textbf{Text-Guided Image Synthesis.} Early works for text-guided image synthesis leverage Generative Adversarial Networks (GAN) \citep{goodfellow2014generative} conditioned on text \citep{tao2022df, xu2018attngan, zhang2021cross, zhu2019dm}.
Based on the success of Transformers \citep{vaswani2017attention}, many subsequent works reframe text-guided image synthesis as a sequence-to-sequence task \citep{zhang2021ernie, ramesh2021zero, ding2021cogview, gafni2022make, yu2022scaling}.
Recently, diffusion models \citep{sohl2015deep, ho2020denoising} gradually become mainstream, which treats image generation as an iterative denoising task.
By injecting text as a condition into the denoising process, many diffusion-based models achieve significant success in text-guided image synthesis \citep{nichol2021glide, ramesh2022hierarchical, saharia2022photorealistic, rombach2022high, balaji2022ediffi, xue2023raphael, feng2023ernie, shen2023advancing, shen2024imagdressing, shen2024boosting, shen2024imagpose, wang2024v, fu2024ap, shen2025long}.
Among them, the Latent Diffusion Model (LDM) \citep{rombach2022high} performs the diffusion and reverse process in the latent space, instead of the pixel space, which largely reduces the computational burden.
Among all the implementations of the Latent Diffusion Model, the stable diffusion model (SD) is the most famous one.
Along with its public availability, the open-source community achieves tremendous success, with the emergence of many excellent fine-tuned models.
These high-quality models are typically fine-tuned on various expert datasets, showcasing diverse denoising capabilities. 
In this paper, we aim to leverage multiple such models to achieve stronger text-to-image generation.

\textbf{Model Merging.} For vision or language understanding tasks, the effectiveness of parameter merging \citep{frankle2020linear, wortsman2022robust, matena2022merging, ilharco2022patching, li2022branch, don2022cold, jin2022dataless} can be interpreted from the perspectives of loss landscape \citep{wortsman2022model, ainsworth2022git, stoica2023zipit} and task arithmetic \citep{ilharco2022editing, ortiz2023task, zhang2023composing}. 
However, merging-based methods have not been widely studied for generation tasks, especially diffusion-based generation.
Recently, many intuitive merging-based methods have emerged.
One of the popular methods is Weighted-Merging, which manually determines weights to merge each U-Net \citep{ronneberger2015u} parameter of multiple SD models.
Although the simplicity, Weighted Merging coarsely allocates weight to all U-Net blocks.
As an improvement, Merge Block Weighted (MBW) allows for manually setting different merging weights for the parameters of distinct U-Net blocks, which provides a more fine-grained merging strategy.
Furthermore, to reduce the reliance on manual weights, autoMBW attempts to automate the merging process, which enumeratively selects the optimal combination of MBW weights by an aesthetic classifier.
However, autoMBW is constrained by the performance bottleneck of the aesthetic classifier, and the enumerative selection leads to a huge time consumption to find the optimal settings.
Such merging methods can be called the static methods.
In this paper, we aim to leverage multiple diffusion models by the dynamic methods, i.e., ensembling.

\textbf{Model Ensembling.} Model ensembling is an effective method to achieve better performance \citep{zhou2012ensemble}, which has been widely applied in various vision understanding tasks, such as classification \citep{zhao2005survey, rokach2010ensemble, yang2010review}, regression \citep{mendes2012ensemble, ren2015ensemble}, clustering \citep{vega2011survey}.
While fewer works focus on the ensembling of generative models, because of the complexity of image space.
Vision-Aided GAN \citep{kumari2022ensembling} focuses on GANs \citep{goodfellow2014generative}, which guides the optimization of a target generator by ensembling pretrained models as a loss.
MagicFusion \citep{zhao2023magicfusion} focuses on the diffusion models, which fuses the predicted noises of two expert U-Net denoisers to implement specific applications, such as style transferring and object binding.
In this paper, we aim to efficiently ensemble multiple diffusion models to achieve general generation improvements\footnote{Note that combining the styles/concepts of multiple diffusion models is also a common goal of ensembling/merging, which is not the interest of our work.}.

\section{Method}

\subsection{Preliminary}

As a type of generative model, the diffusion model \citep{sohl2015deep,ho2020denoising} consists of two processes, which are the diffusion process and the reverse process, respectively.
In the diffusion process (i.e., the forward process), the Gaussian noises are iteratively added to degrade the images over $T$ steps until the images become completely random noise.
In the reverse process, a trained denoiser is used to iteratively generate images from the sampled Gaussian noise.

When training, given an input image $\mathbf{x}_0$ and the additional condition $\mathbf{c}$ (e.g., encoded textual prompts etc.), the denoising loss is defined as
\begin{equation}
    \label{eq:denoising-loss}
    \mathcal{L}_\text{denoise} = \mathbb{E}_{\mathbf{x}_0, \boldsymbol{\epsilon}\sim\mathcal{N}(\mathbf{0}, \mathbf{I}), \mathbf{c}, t}
    \left\Vert
    \boldsymbol{\epsilon}_\theta\left(\mathbf{x}_t, \mathbf{c}, t\right)
    - \boldsymbol{\epsilon}
    \right\Vert^2 \ .
\end{equation}
Among them, $\mathbf{x}_t=\sqrt{\alpha_t}\mathbf{x}_0+\sqrt{1-\alpha_t}\boldsymbol{\epsilon}$ is the noisy image at timestep $t\in[1,T]$, where $\alpha_t$ is a predefined scalar from the noise scheduler. 
$\boldsymbol{\epsilon}$ is the added noise. 
$\boldsymbol{\epsilon}_\theta$ is the denoiser with the learnable parameters $\theta$, which predicts the noise to be removed from the noisy image.

LDM stands out as one of the most popular diffusion models. It performs the diffusion and reverse process in the latent space which is encoded by a Variational Auto-Encoder (VAE) \citep{kingma2013auto, rezende2014stochastic}.
In LDM, the U-Net structure \citep{ronneberger2015u} is used for denoiser, which contains three parts of blocks, which are the down-sampling blocks, the middle block, and the up-sampling blocks, respectively.
Each down-sampling block is skip-connected with a corresponding up-sampling block.
These blocks are composed of Transformer \citep{vaswani2017attention} layers and ResNet \citep{he2016deep} layers.
In the Transformer layers, the cross-attention mechanism is employed to incorporate conditional textual prompts, which plays a crucial role in text-guided image generation.

\begin{figure}[t]
    \centering
    \includegraphics[width=1\columnwidth]{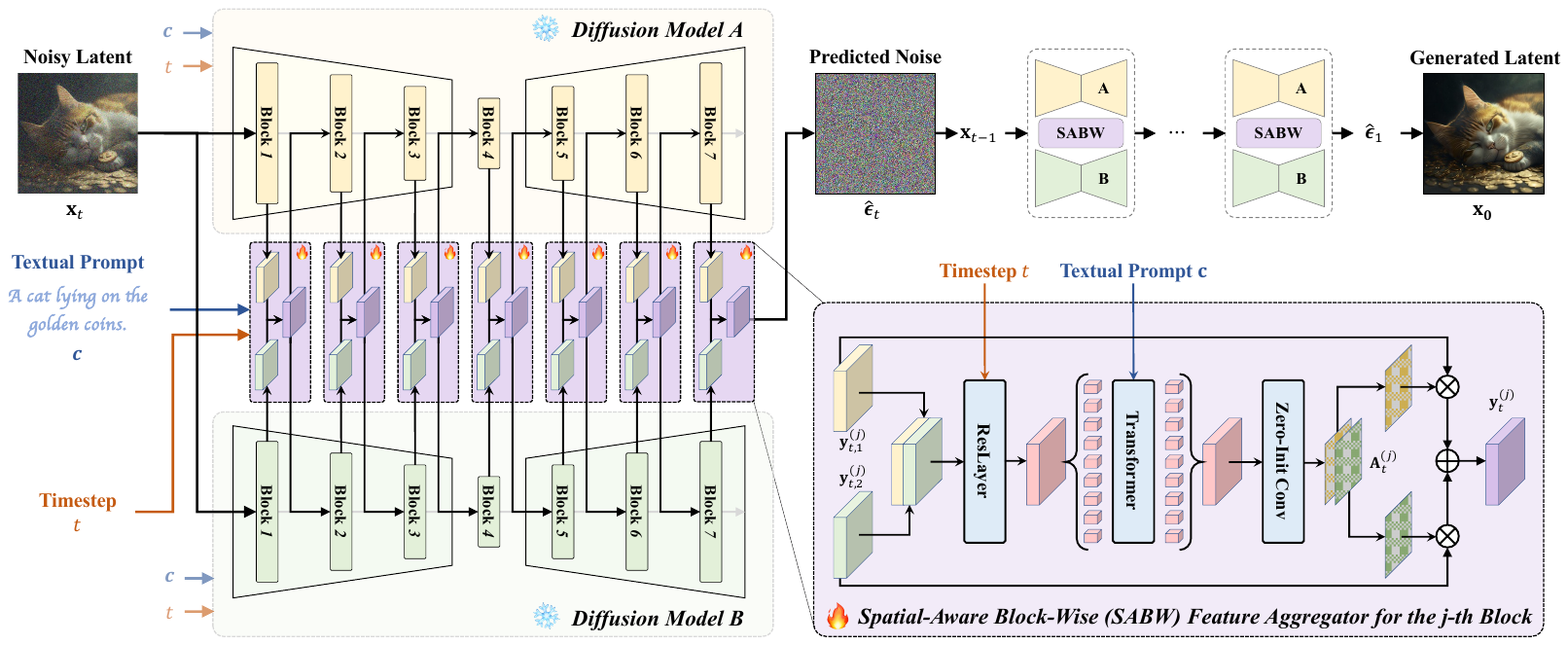}
    \vspace{-5pt}
    \caption{Framework of ensembling multiple diffusion models by our AFA method.}
    \vspace{-5pt}
    \label{fig:framework}
\end{figure}

\subsection{Adaptive Feature Aggregation}

Our proposed Adaptive Feature Aggregation (AFA) ensembles multiple diffusion models that share the same architecture but different parameters.
Specifically, SABW is designed to integrate intermediate features from multiple U-Net denoisers at the block level. 
Figure~\ref{fig:framework} demonstrates an example of ensembling two diffusion models, and each U-Net denoiser contains seven blocks by SABW.
The forward process of AFA can be seen in Appendix~\ref{app:algo}.

Given $N$ diffusion models to be ensembled, whose U-Net denoisers contain $K$ blocks, 
let $\boldsymbol{\epsilon}_{\theta_i}$ be the $i$-th U-Net denoiser, and $\boldsymbol{\epsilon}_{\theta_i}^{(j)}$ be its $j$-th block.
For the timestep $t$, it transforms the input features $\mathbf{x}_t^{(j)}$ into the output feature by 
\begin{equation}
    \label{eq:unet-denoiser}
    \mathbf{y}_{t,i}^{(j)}=\boldsymbol{\epsilon}_{\theta_i}^{(j)}(\mathbf{x}_t^{(j)}, \mathbf{c}, t) \ .
\end{equation}
Note that because of the same architecture of all $\boldsymbol{\epsilon}_{\theta_i}$, $\mathbf{y}_{t, i}^{(j)}$ has the same shape, i.e., $\mathbf{y}_{t, i}^{(j)}\in\mathbb{R}^{h_j\times w_j\times c_j}$, where $h_j$, $w_j$, and $c_j$ are the height, width, and channels, respectively.
To achieve the output feature for the next block, a block $f_{\varphi}^{(j)}$ of SABW is employed to aggregate $\{\mathbf{y}_{t, i}^{(j)}\}_{i=1}^N$,
\begin{equation}
    \label{eq:ba-adpter}
    \mathbf{y}_t^{(j)}=f_{\varphi}^{(j)}(\{\mathbf{y}_{t,i}^{(j)}\}_{i=1}^N, \mathbf{c}, t) \in \mathbb{R}^{h_j\times w_j\times c_j} \ ,
\end{equation}
where $\varphi$ is the parameters of the SABW block.
As a whole, the $j$-th ensembled block of these U-Net denoisers can be formulated as
\begin{equation}\label{eq:ensembled-block}
    \mathbf{y}_t^{(j)}=\mathcal{F}_{\Theta}^{(j)}(\mathbf{x}_t^{(j)}, \mathbf{c}, t) \ ,
\end{equation}
where $\Theta$ is the parameter collection of $\varphi$ and all the $\theta_i$.
It can be deduced that an ensembled block operates with the same functionality as a single block.

\subsection{Spatial-Aware Block-Wise Feature Aggregator}

The Spatial-Aware Block-Wise (SABW) feature aggregator is learned to aggregate output features of each block from multiple U-Net denoisers. 
We design SABW to learn spatial attention to reweight the contributions of these U-Net denoisers at each spatial position.
Specifically, the spatial attention map $\mathbf{A}_t^{(j)}$ for $N$ U-Net denoisers is achieved by
\begin{align}
    \label{eq:spatial-attention-map}
    \mathbf{F}_t^{(j)} &= g_{\varphi}^{(j)}(\{\mathbf{y}_{t,i}^{(j)}\}_{i=1}^N, \mathbf{c}, t)\in\mathbb{R}^{h_j\times w_j\times N} \ , \\
    \mathbf{A}_t^{(j)} &= \text{softmax}(\mathbf{F}_t^{(j)})\in[0,1]^{h_j\times w_j\times N} \ ,
\end{align}
where, $\mathbf{F}_t^{(j)}$ is the output of the learnable network of SABW $g_\varphi$.
The ensembled output feature is
\begin{equation}
    \mathbf{y}_t^{(j)} = 
    \mathcal{F}_{\Theta}^{(j)}(\mathbf{x}_t^{(j)}, \mathbf{c}, t) = 
    \sum_{i=1}^N\mathbf{A}_{t,i}^{(j)} \otimes \mathbf{y}_{t,i}^{(j)} \ ,
\end{equation}
where $\otimes$ is the element-wise multiplication.
$\mathbf{A}_{t,i}^{(j)}\in\mathbb{R}^{h_j\times w_j\times 1}$ is the spatial attention map for the $i$-th U-Net denoiser.

Note that the spatial-attention mechanism of SABW differs to the typical multi-head self-attention mechanism \citep{vaswani2017attention}, which simply uses three linear layers to project the input features into three separate spaces (i.e., query, key, and value), and then computes the attention map across all the projected features.
Our spatial-attention mechanism is based on an important experimental insight as shown in Figure~\ref{fig:positional-denoising-capability}.
The divergent denoising capabilities of multiple diffusion models are influenced by the prompts and the denoising steps, and they dynamically vary across different spatial locations.
Thus, the projection method needs to account for the input prompt and the current denoising step.
And it is sufficient to compute the attention map on each spatial location individually. 
As a result, better generation performance can be achieved by enabling multiple models to collaborate at different spatial locations under the conditions of the input prompt and the current denoising step.

Specifically, based on the denoising loss (i.e., Eq.~\ref{eq:denoising-loss}), the denoising capability\footnote{Compared with the denoising loss in Eq.~\ref{eq:denoising-loss}, the denoising capability is defined by adding the negative sign, because of a negative correlation between the denoising capability and the denoising loss.} of $\boldsymbol{\epsilon}_\theta$ for $(\mathbf{x}_0, \mathbf{c})$ at timestep $t$ can be defined as $-\mathbb{E}_{\boldsymbol{\epsilon}\sim\mathcal{N}(\mathbf{0}, \mathbf{I})}\Vert \hat{\boldsymbol{\epsilon}}_t - \boldsymbol{\epsilon} \Vert^2$, where $\hat{\boldsymbol{\epsilon}}_t = \boldsymbol{\epsilon}_\theta(\mathbf{x}_t,\mathbf{c}, t)$.
For the spatial location $(x,y)$, the denoising capability is formulated as $-\mathbb{E}_{\boldsymbol{\epsilon}\sim\mathcal{N}(\mathbf{0}, \mathbf{I})}\Vert \hat{\boldsymbol{\epsilon}}_{(x,y)} -\boldsymbol{\epsilon}_{(x,y)} \Vert^2$, which considers the noise prediction errors at the spatial location $(x,y)$.

For example, as shown in Figure~\ref{fig:positional-denoising-capability}, we employ three SD models to present their positional denoising capability on the specific image and textual prompt by Monte-Carlo sampled noises.
We demonstrate the results of four patches which indicate that the divergent denoising capabilities of the three models vary at different spatial locations.
SABW is learned to incorporate these denoisers to achieve stronger denoising capability at all spatial locations at the block level.

To implement $g_{\varphi}$ in Eq.~\ref{eq:spatial-attention-map}, SABW applies a ResNet \citep{he2016deep} layer to introduce timesteps by adding the time embedding into the feature map, and a Transformer \citep{vaswani2017attention} layer to introduce textual prompts by the cross-attention,
\begin{align}
    \mathbf{Y}_t^{(j)} &\leftarrow \text{concat}(\{\mathbf{y}_{t,i}^{(j)}\}_{i=1}^N)\in\mathbb{R}^{h_j\times w_j\times (N\cdot c_j)} \ , \\
    \mathbf{H}_t^{(j)} &\leftarrow \text{ResLayer}(\mathbf{Y}_t^{(j)}+\gamma(t))\in\mathbb{R}^{h_j\times w_j\times d} \ , \\
    \mathbf{O}_t^{(j)} &\leftarrow \text{TransformerLayer}(\mathbf{H}_t^{(j)},\mathbf{c})\in\mathbb{R}^{h_j\times w_j\times d} \ , \\
    \mathbf{F}_t^{(j)} &\leftarrow \text{ZeroInitConv}(\mathbf{O}_t^{(j)})\in\mathbb{R}^{h_j\times w_j\times N} \ ,
\end{align}
where $\gamma$ is used to map the timestep into the time embedding with the same shape of $\mathbf{Y}_t^{(j)}$. 
$d$ is the hidden state dimension. $\text{ZeroInitConv}(\cdot)$ is the convolution layer initialized by zero, which leads to equal attention weights for all denoisers before training. We aspire for the model to commence training from a completely equitable state. 

\begin{table}[t]
    \centering
    \begin{minipage}{0.48\textwidth}\setlength{\tabcolsep}{1.45pt}
    \centering
    {\tiny\begin{tabular}{lcccccccc}
    \toprule
    & \multicolumn{4}{c}{COCO 2017} & \multicolumn{4}{c}{Draw Bench Prompts} \\ \cmidrule(lr){2-5} \cmidrule(lr){6-9}
    & FID $\downarrow$ & IS & CLIP-I & CLIP-T & AES & PS & HPSv2 & IR \\ \midrule
    Base Model A & 
    13.01 & 5.65 & .6724 & .2609 & 5.4102 & 21.6279 & 27.8007 & .3544 \\
    Base Model B & 
    13.45 & 5.43 & .6775 & .2652 & 5.5013 & 21.4624 & 27.7246 & .2835 \\
    Base Model C & 
    12.32 & 6.32 & .6890 & .2566 & 5.4881 & 21.8031 & 27.9652 & .3922 \\ \midrule \midrule
    Wtd. Merging &
    10.65 & 6.93 & .6861 & .2626 & 5.4815 & 21.7272 & 27.9086 & .3909 \\
    MBW &
    11.03 & 6.51 & .6870 & .2624 & 5.4812 & 21.7201 & 27.9080 & .3922 \\
    autoMBW &
    13.35 & 5.51 & .6772 & .2577 & 5.5056 & 21.4785 & 27.8192 & .3672 \\ \midrule
    MagicFusion & 
    10.53 & 6.85 & .6751 & .2620 & 5.3431 & 21.3840 & 27.8105 & .3317 \\
    \textbf{AFA (Ours)} &
    \textbf{9.76} & \textbf{7.14} & \textbf{.6926} & \textbf{.2675} & \textbf{5.5201} & \textbf{21.8263} & \textbf{27.9734} & \textbf{.4388} \\
    \bottomrule
    \end{tabular}}
    \vspace{-5pt}
    \caption{Quantitative comparison for the base models in Group I (i.e., ER, MMR, and RV).}
    \vspace{-5pt}
    \label{tab:performance-group-1}
    \end{minipage}
    \hfill
    \begin{minipage}{0.48\textwidth}\setlength{\tabcolsep}{1.45pt}
    \centering
    {\tiny \begin{tabular}{lcccccccc} 
    \toprule
    & \multicolumn{4}{c}{COCO 2017} & \multicolumn{4}{c}{Draw Bench Prompts} \\ \cmidrule(lr){2-5} \cmidrule(lr){6-9}
    & FID $\downarrow$ & IS & CLIP-I & CLIP-T & AES & PS & HPSv2 & IR \\ \midrule
    Base Model A & 
    12.12 & 5.66 & .6849 & .2623 & 5.5641 & 21.8013 & 28.0183 & .4238 \\
    Base Model B & 
    12.41 & 5.59 & .6818 & .2580 & 5.5027 & 21.7249 & 28.0343 & .4202 \\
    Base Model C & 
    12.05 & 5.95 & .6638 & .2642 & 5.5712 & 21.4936 & 27.8089 & .3367 \\ \midrule \midrule
    Wtd. Merging &
    11.53 & 6.56 & .6824 & .2631 & 5.5756 & 21.7516 & 28.0014 & .4387 \\
    MBW &
    12.06 & 6.42 & .6826 & .2632 & 5.5772 & 21.7487 & 28.0029 & .4396 \\
    autoMBW &
    12.39 & 5.62 & .6774 & .2588 & 5.5478 & 21.5135 & 27.9873 & .3513 \\ \midrule
    MagicFusion & 
    11.63 & 7.13 & .6790 & .2640 & 5.4674 & 21.4270 & 27.9608 & .4194 \\
    \textbf{AFA (Ours)} &
    \textbf{10.27} & \textbf{7.42} & \textbf{.6855} & \textbf{.2717} & \textbf{5.5798} & \textbf{21.8059} & \textbf{28.0371} & \textbf{.4892} \\
    \bottomrule
    \end{tabular}}
    \vspace{-5pt}
    \caption{Quantitative comparison for the base models in Group II (i.e., AR, CR, and RCR).}
    \vspace{-5pt}
    \label{tab:performance-group-2}
    \end{minipage}
\end{table}

\subsection{Training}

Given the ensembled denoiser $\mathcal{F}_\Theta$, the training object is same as the denoising loss in Eq.~\ref{eq:denoising-loss},
\begin{equation}
    \mathcal{L} = \mathbb{E}_{\mathbf{x}_0, \boldsymbol{\epsilon}\sim\mathcal{N}(\mathbf{0}, \mathbf{I}), \mathbf{c}, t}
    \left\Vert
    \mathcal{F}_\Theta\left(\mathbf{x}_t, \mathbf{c}, t\right)
    -\boldsymbol{\epsilon}
    \right\Vert^2 \ .
\end{equation}
Note that the parameters $\Theta$ contains both the parameters of SABW $\varphi$ and the parameters of all U-Net denoisers $\{\theta_i\}_{i=1}^N$, where $\varphi$ are learnable, and $\{\theta_i\}_{i=1}^N$ are frozen.

A well-trained AFA ensembles multiple U-Net denoisers to enhance the denoising capability in every block when experiencing different prompts, initial noises, denoising steps, and spatial locations.
It leads to smaller denoising errors and stronger generation performance.

\subsection{Inference}

The inference of AFA starts from the sampled Gaussian noise.
Then, the diffusion scheduler (e.g., DDIM \citep{song2020denoising}, PNDM \citep{liu2022pseudo}, DPM-Solver \citep{lu2022dpm, lu2022dpm2}, 
etc.) is applied to generate images with multiple denoising steps.

For each inference step, the noise prediction relies on the technique of Classifier-Free Guidance (CFG) \citep{ho2022classifier}, which is formulated as $\hat{\boldsymbol{\epsilon}}_t^\text{pred} = \hat{\boldsymbol{\epsilon}}_t^\text{uc}+\beta_\text{CFG}(\hat{\boldsymbol{\epsilon}}_t^\text{c}-\hat{\boldsymbol{\epsilon}}_t^\text{uc})$ .
Among them, $\hat{\boldsymbol{\epsilon}}_t^\text{pred}$, $\hat{\boldsymbol{\epsilon}}_t^\text{uc}$, $\hat{\boldsymbol{\epsilon}}_t^\text{c}$, and $\beta_\text{CFG}$ are the predicted noise, the predicted noise with condition, the predicted noise without condition, and a guidance scale, respectively. 
The latent for the next step is $\mathbf{x}_{t-1}=\frac{1}{\sqrt{\alpha_t}}\mathbf{x}_{t}-\frac{\sqrt{1-\alpha_t}}{\sqrt{\alpha_t}}\hat{\boldsymbol{\epsilon}}_t^\text{pred}$ .
Finally, the generated image is achieved from the latent by a VAE decoder.

\section{Experiments}

\subsection{Experiment Settings}

\textbf{Base Models.} We select six popular models with the same architecture from SDv1.5\footnote{\url{https://huggingface.co/stable-diffusion-v1-5/stable-diffusion-v1-5}} in CivitAI.
There are 12 down-sampling blocks, 1 middle block, and 12 up-sampling blocks for each model.
These models are randomly divided into two groups, with three models in each group. 
The models from the same group will be ensembled.
\begin{itemize}[leftmargin=1em]
    \item \textbf{Group I} contains EpicRealism\footnote{\url{https://civitai.com/models/25694?modelVersionId=134065}} (ER), MagicMixRealistic\footnote{\url{https://civitai.com/models/43331?modelVersionId=176425}} (MMR), and RealisticVision\footnote{\url{https://civitai.com/models/4201?modelVersionId=130072}} (RV);
    \item \textbf{Group II} contains AbsoluteReality\footnote{\url{https://civitai.com/models/81458?modelVersionId=108576}} (AR), CyberRealistic\footnote{\url{https://civitai.com/models/15003?modelVersionId=89680}} (CR), and RealCartoonReslistic\footnote{\url{https://civitai.com/models/97744?modelVersionId=104496}} (RCR).
\end{itemize}

\textbf{Training.} AdamW \citep{loshchilov2017decoupled} is used as the optimizer with a learning rate of 0.0001 and a weight decay of 0.01.
Note that only the parameters of SABW are optimized, while those of U-Net denoisers are frozen.
Our AFA framework is trained on 10,000 samples from the dataset JourneyDB \citep{pan2023journeydb} for 10 epochs with batch size 8.
To enable CFG, we use a probability of 0.1 to drop textual prompts.

\textbf{Evaluation Protocols.} For a fair comparison, all the methods generate 4 images by DDIM \cite{ho2020denoising} for 50 inference steps.
The CFG weight $\beta_\text{CFG}$ is set to 7.5.
We evaluate the models with two datasets, which are COCO 2017 \citep{lin2014microsoft} and Draw Bench Prompts \citep{saharia2022photorealistic}, respectively.
For COCO 2017, we apply four metrics, which are Fréchet Inception Distance (FID), Inception Score (IS), CLIP-I, and CLIP-T, respectively.
For Draw Bench Prompts,we apply 4 evaluation metrics, which are AES\footnote{\url{https://github.com/christophschuhmann/improved-aesthetic-predictor}}, Pick Score (PS) \citep{kirstain2023pick}, HPSv2 \citep{wu2023human}, and Image Reward (IR) \citep{xu2023imagereward}, respectively. 
More details can be found in Appendix~\ref{app:details-about-evaluation-protocols}.

\textbf{Baselines.} We compare our AFA with several methods, including three merging-based methods (i.e., Weighted-Merging, MBW, and autoMBW) and one ensembling-based method (i.e., MagicFusion \citep{zhao2023magicfusion}).
The evaluation of these baselines follows the protocols.


\begin{figure}[t]
    \centering
    \begin{minipage}[t]{0.48\linewidth}
    \includegraphics[width=1\columnwidth]{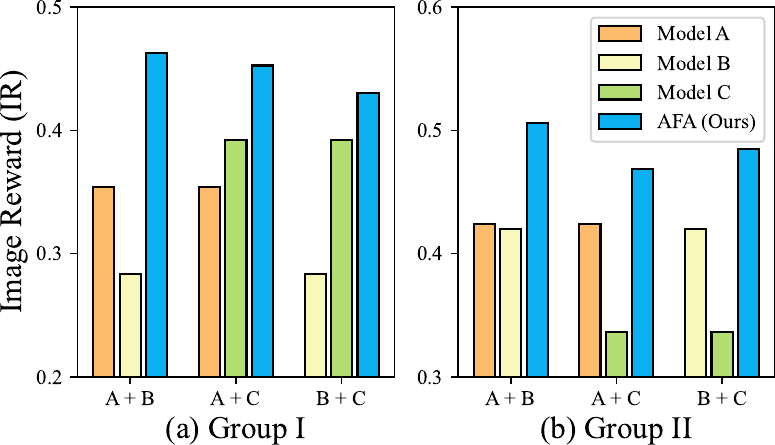}
    \vspace{-5pt}
    \caption{Quantitative comparison between AFA with the two base models.}
    \vspace{-5pt}
    \label{fig:ensembling-two-models}
    \end{minipage}
    \quad
    \begin{minipage}[t]{0.48\linewidth}
    \centering
    \includegraphics[width=1\columnwidth]{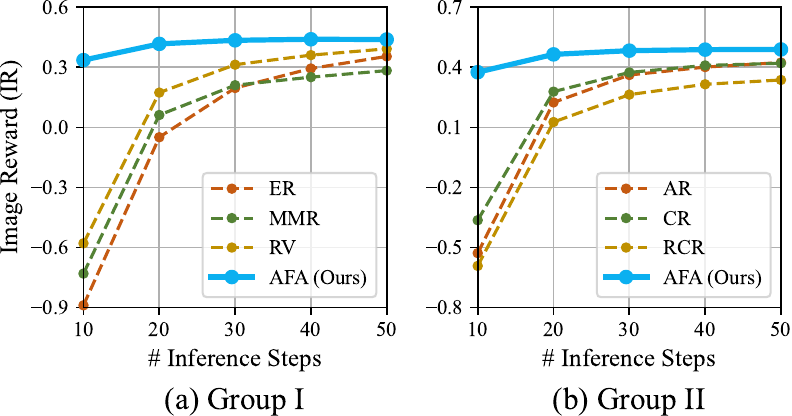}
    \vspace{-5pt}
    \caption{Effect of varying inference steps.}
    \vspace{-5pt}
    \label{fig:inference-steps}
    \end{minipage}
\end{figure}

\subsection{Quantitative Comparison}

The results of the two evaluation protocols are summarized in Tables~\ref{tab:performance-group-1}~and~\ref{tab:performance-group-2}.
Generally, compared with base models, performance improvements are achieved by our AFA.
For COCO 2017, after being ensembled by AFA, FID, IS, CLIP-I, and CLIP-T exhibit enhancements under both Group I and Group II.
This indicates that the quality of our generated images is more closely with the dataset images. 
Additionally, the contexts of our generated images align more closely with those of the dataset images and corresponding captions.
For Draw Bench Prompts, all four metrics show improvements under both Group I and Group II.
The improvement on AES suggests that the aesthetics of our generated images surpass those generated by the base models.
The improvements on PS, HPSv2, and IR suggest that the general generation capability can be boosted by applying AFA.
Overall, the quantitative results validate the effectiveness of our AFA in ensembling diffusion models.

While MagicFusion attains impressive performance in style-transferring and object-binding by ensembling two U-Net denoisers, it fails to achieve superior general generation when ensembling multiple U-Net denoisers.
The merging-based methods, such as Weighted Merging, MBW, and autoMBW, do not always produce models with superior generation capabilities.
For instance, under Group II, Weighted Merging achieves a better IR compared to the base models, but under Group I, it yields a lower IR.
This could be attributed to the fixed contributions of merged models, which are determined by a set of predefined merging weights.
Consequently, these methods fail to effectively facilitate collaboration between models to compensate for each other's weaknesses and improve their generative capabilities.
Compared with these baselines, our AFA consistently achieves better performance, demonstrating its superiority.

As depicted in Figure~\ref{fig:ensembling-two-models}, we use AFA to ensemble two models from Group I or Group II, and present the IR scores of base models and the ensembled models on Draw Bench Prompts.
The results show that AFA significantly improves the performance when ensembling two U-Net denoisers.
The results presented in Table~\ref{tab:performance-group-1}, Table~\ref{tab:performance-group-2}, and Figure~\ref{fig:ensembling-two-models} indicate that AFA can be effectively applied to various model combinations and can be extended to accommodate different numbers of models.
Meanwhile, it consistently achieves superior generation performance compared to a single base model.

\begin{figure}[t]
    \centering
    \includegraphics[width=1\columnwidth]{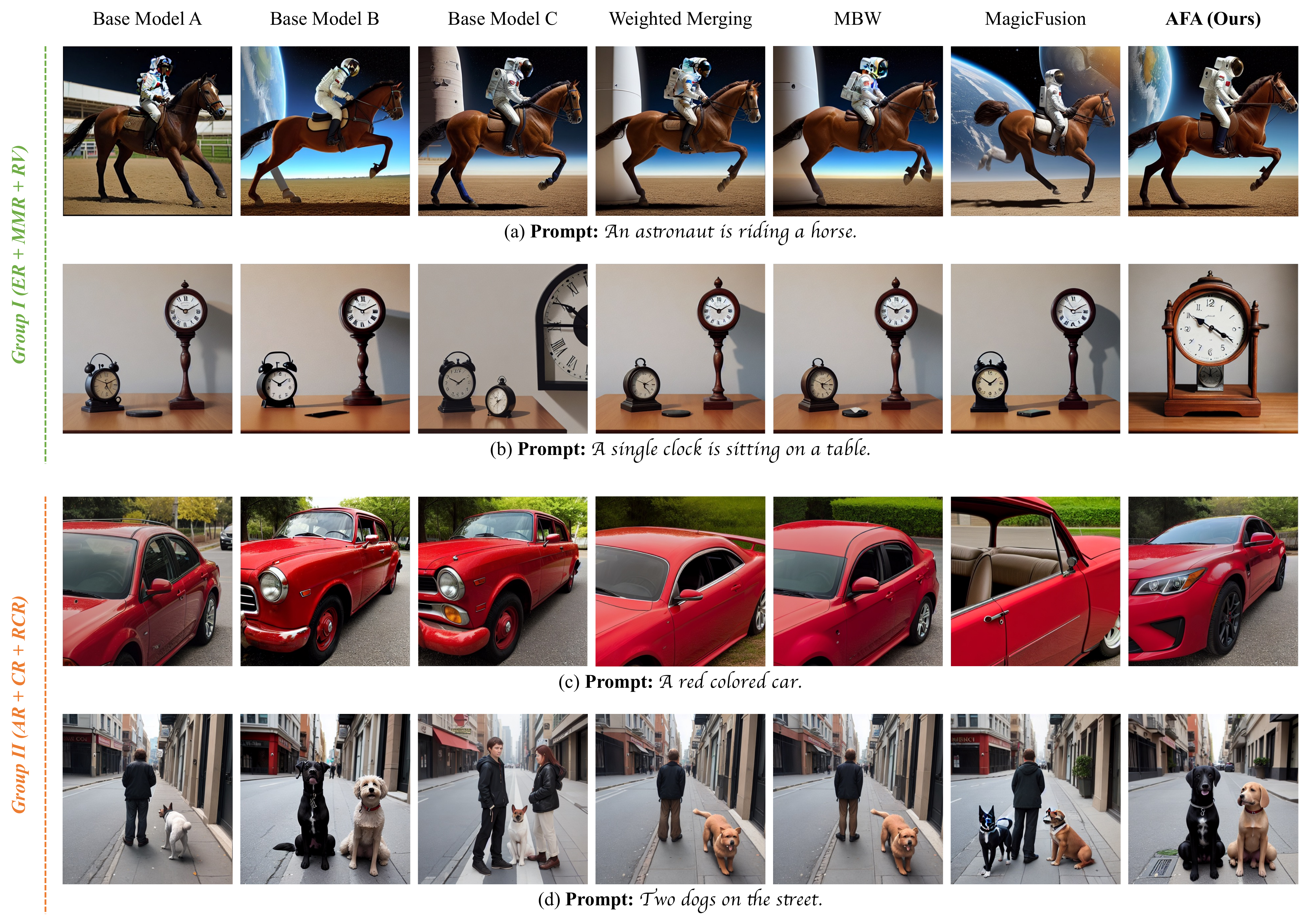}
    \vspace{-5pt}
    \caption{Qualitative comparison between AFA with the base models and the baselines.}
    \vspace{-5pt}
    \label{fig:qualitative-comparison}
\end{figure}

\subsection{Qualitative Comparison}

As shown in Figure~\ref{fig:qualitative-comparison}, we present a qualitative comparison of the results from the base models, some of the baseline methods, and our AFA, which allows us to make several insightful observations.
Firstly, AFA can generate images with better aesthetics.
For example, in Figure~\ref{fig:qualitative-comparison}~(a) and (c), AFA can generate images with improved composition and finer details, compared to both the base models and the baseline methods.
Secondly, AFA excels in achieving superior context alignment.
For example, in Figure~\ref{fig:qualitative-comparison}~(b), all the base models and the baseline methods generate images containing more than \textit{one clock}, despite the textual prompt specifying \textit{a single clock}. 
In contrast, only our AFA generates the image with just one clock, accurately reflecting the provided context.
Thirdly, AFA can focus on the correct context of the base models and drop out the incorrect context.
For example, in Figure~\ref{fig:qualitative-comparison}~(d), only the base model B generates the image with \textit{two dogs}, while the baseline methods do not fully trust the base model B, and generate the images with \textit{people}.
Only our AFA fully trusts the base model B, and generates the image with \textit{two dogs}, which aligns with the textual context.

\begin{figure}[t]
    \centering
    \includegraphics[width=\columnwidth]{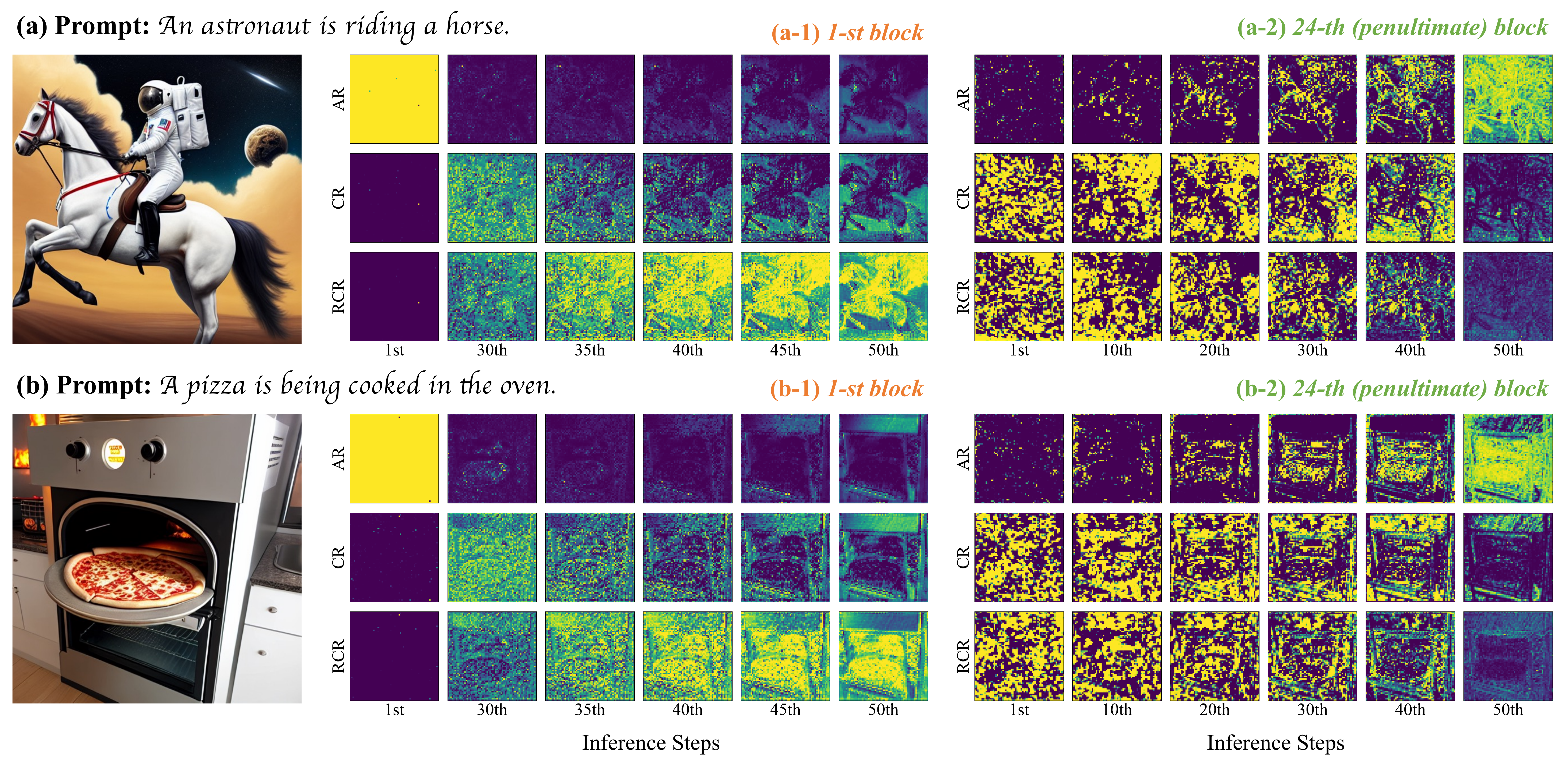}
    \vspace{-5pt}
    \caption{Visualization of the learned attention maps. 
    In each attention map, lighter-coloered positions mean higher attention weights, while darker-coloered positions represent lower weights.}
    \vspace{-5pt}
    \label{fig:attn-maps}
\end{figure}

\subsection{Model Study}
\label{sec:model-study}

\textbf{Visualization of Attention Maps.} As shown in Figure~\ref{fig:attn-maps}, we select to visualize the attention maps of the first and the penultimate (i.e., 24-th) blocks, both of which are learned by ensembling models in Group II through our AFA.
For the first block, the learned attention maps seem to concentrate solely on the feature map of AR at the first inference stage.
However, at the last few inference steps (i.e., from the 30th to the 50th step), the roles of the feature maps from different models begin to diverge. 
The attention maps tend to highlight the contextual body (i.e., the \textit{astronaut} and the \textit{horse} in Figure~\ref{fig:attn-maps}~(a), and the \textit{pizza} and the \textit{oven} in Figure~\ref{fig:attn-maps}~(b)) in the feature map of RCR, while focusing on the background elements in the feature maps of AR and CR.
For the penultimate block, as the inference progresses, the learned attention maps gradually emphasize the contextual body within the feature map of AR, while concentrating on the background within the feature maps of both CR and RCR.
Overall, the visualization of the learned attention maps indicates that our AFA can effectively ensemble diffusion models based on the context and the timesteps.



\begin{figure}[t]
    \centering
    \includegraphics[width=1\columnwidth]{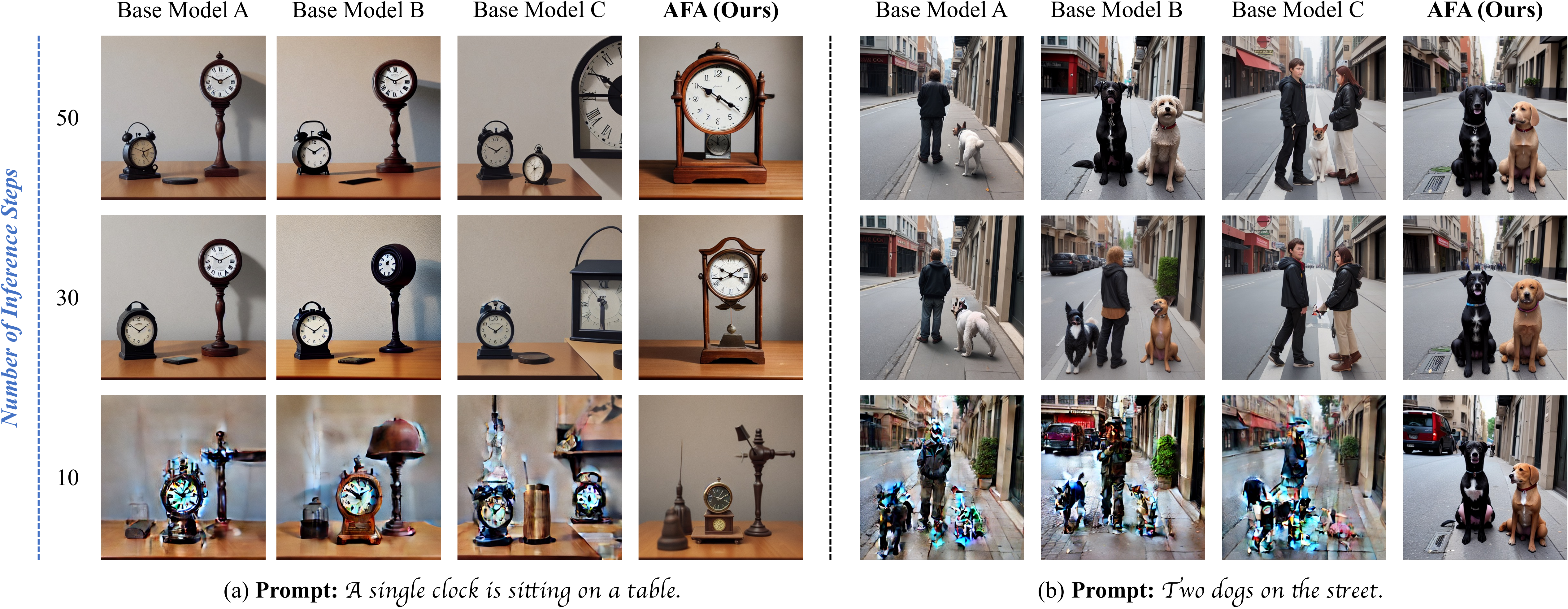}
    \vspace{-5pt}
    \caption{Qualitative comparison under different inference steps.}
    \vspace{-5pt}
    \label{fig:qualitative-inference-steps}
\end{figure}

\textbf{Effect of Fewer Inference Steps.} As shown in Figure~\ref{fig:inference-steps} and Figure~\ref{fig:qualitative-inference-steps}, we investigate the resilience of our AFA to a reduction in the inference steps.
Figure~\ref{fig:inference-steps} clearly shows a significant drop in the performance of the base models when the number of inference steps is reduced.
However, the performance of our AFA remains almost stable when the number of inference steps is reduced from 50 to 20, and only experiences a slight decline when the number of inference steps is reduced to 10.
Additionally, in Figure~\ref{fig:qualitative-inference-steps}, we can find that the quality of the images generated by the base models deteriorates with the reduction of inference steps.
In contrast, our AFA stably generates high-quality images.
Overall, it indicates that greater tolerance for fewer inference steps will be achieved after ensembling by AFA. 
Furthermore, while a single inference step may require a proportional increase in time, the total time consumed by the entire inference process does not escalate in the same manner.

\begin{wraptable}{r}{0.55\textwidth}\scriptsize \setlength{\tabcolsep}{9.16pt}
  \centering
  \begin{tabular}{lcc}
    \toprule
    & IR (Group I) & IR (Group II) \\ \midrule
    \textbf{AFA (Full Model)}    & \textbf{.4388} & \textbf{.4892} \\ \midrule
    Only Ensembling Last Block   & .4176          & .4374          \\
    Block-Wise Averaging         & .4001          & .4372          \\
    Noise Averaging              & .3919          & .4355          \\ \midrule
    w/o Spatial Location         & .4044          & .4610          \\
    w/o Timestep                 & .4235          & .4622          \\
    w/o Textual Condition        & .4163          & .4559          \\
    \bottomrule
  \end{tabular}
  \caption{Ablation study of AFA.}
  \label{tab:ablation-study}
\end{wraptable}

\textbf{Ablation Study.} As shown in Table~\ref{tab:ablation-study}, we conduct the ablation study to investigate the necessity of each component within our AFA.
Firstly, we conduct experiments with the following settings.
(1) \textit{\underline{Only Ensembling Last Block}}: The AFA ensembles only the last block, leaving all other blocks unaltered.
(2) \textit{\underline{Block-Wise Averaging}}: Each block is ensembled by averaging the output feature maps, rather than using SABW.
(3) \textit{\underline{Noise Averaging}}: Only the last block is ensembled by averaging the output feature maps (i.e., predicted noises).
The observed performance declines in these settings suggest that the design of the AFA effectively contributes to the ensembling of diffusion models, thereby leading to improved generation.
Next, we conduct an ablation study on the spatial location from AFA, which learn only a single attention score (i.e., a scalar) for the feature of each U-Net denoiser, as opposed to an attention map.
The observed decrease in performance suggests that learning a spatial attention map to adaptively adjust the weights of various U-Net denoisers in the level of spatial location is beneficial for the ensemble of different diffusion models. 
Finally, we individually ablate the introduced timestep and the textual condition.
The decreasing performance indicates that both the timestep and textual condition play crucial roles for AFA.

\textbf{Discussion on Computational Efficiency.} (1) \textit{\underline{Ensembling Methods.}} Compared with another ensembling method (i.e., MagicFusion \citep{zhao2023magicfusion}), the computational efficiency of our AFA for a single inference step aligns closely with it.
Both AFA and MagicFusion necessitate running all base models in one inference step.
Although SABW in AFA introduces additional parameters, which leads to more computations, it has fewer than the base models (about 1/16).
To be precise, our AFA may be marginally less efficient than MagicFusion on single inference steps, but the difference is practically negligible. 
(2) \textit{\underline{Merging Methods.}} Because the various models are collapsed into one model, the computational efficiency of the merging methods equals that of a single base model.
For the base models and the merging methods, a single image generation takes about 7 seconds with 50 inference steps.
When ensembling three such base models, a single generation will take about 22 seconds with the same inference steps.
Compared with ensembling, the computation efficiency of one inference step of the merging methods is significantly higher.
However, as we discussed in \textit{Effect of Fewer Inference Steps} of Section~\ref{sec:model-study}, AFA shows remarkable tolerance for fewer inference steps.
When limited to 20 inference steps, the performance of AFA does not significantly decline (as shown in Figure~\ref{fig:inference-steps}), while the time consumed is reduced to about 8.8 seconds.
This suggests that AFA has comparable computational efficiency with that of the base models or the merging methods throughout the entire inference process.

\section{Conclusion}

In this paper, we aim to ensemble multiple diffusion models to achieve better generation performance.
To this end, we propose the Adaptive Feature Aggregation (AFA) method, which dynamically adjusts the contributions of multiple models at the feature level according to different prompts, initial noises, denoising steps, and spatial locations.
Specifically, we design a lightweight Spatial-Aware Block-Wise feature aggregator that produces attention maps to adaptively aggregate the block-level intermediate features from multiple U-Net denoisers. 
The quantitative and qualitative experiments demonstrate that AFA achieves improvements in image quality and context alignment.

\textbf{Limitation.} A significant limitation of our AFA is that a single inference step may necessitate a proportional increase in inference time, resulting in lower computational efficiency for that step. 
Fortunately, due to its high tolerance for fewer inference steps, the total time consumed by the entire inference process does not escalate in the same manner.
This results in a computational efficiency that is on par with that of the base models or the merging methods.

\section{Acknowledgments}

This work is supported by the National Natural Science Foundation of China under Grants Nos. 62441225, 61972192, 62172208, 61906085. 
This work is partially supported by Collaborative Innovation Center of Novel Software Technology and Industrialization. 
This work is supported by the Fundamental Research Funds for the Central Universities under Grant No.14380001.

\bibliography{iclr2025_conference}

\begin{thebibliography}{78}
\providecommand{\natexlab}[1]{#1}
\providecommand{\url}[1]{\texttt{#1}}
\expandafter\ifx\csname urlstyle\endcsname\relax
  \providecommand{\doi}[1]{doi: #1}\else
  \providecommand{\doi}{doi: \begingroup \urlstyle{rm}\Url}\fi

\bibitem[Ainsworth et~al.(2022)Ainsworth, Hayase, and
  Srinivasa]{ainsworth2022git}
Samuel~K Ainsworth, Jonathan Hayase, and Siddhartha Srinivasa.
\newblock Git re-basin: Merging models modulo permutation symmetries.
\newblock \emph{arXiv preprint arXiv:2209.04836}, 2022.

\bibitem[Balaji et~al.(2022)Balaji, Nah, Huang, Vahdat, Song, Kreis, Aittala,
  Aila, Laine, Catanzaro, et~al.]{balaji2022ediffi}
Yogesh Balaji, Seungjun Nah, Xun Huang, Arash Vahdat, Jiaming Song, Karsten
  Kreis, Miika Aittala, Timo Aila, Samuli Laine, Bryan Catanzaro, et~al.
\newblock ediffi: Text-to-image diffusion models with an ensemble of expert
  denoisers.
\newblock \emph{arXiv preprint arXiv:2211.01324}, 2022.

\bibitem[Chi et~al.(2022)Chi, Dong, Huang, Dai, Ma, Patra, Singhal, Bajaj,
  Song, Mao, et~al.]{chi2022representation}
Zewen Chi, Li~Dong, Shaohan Huang, Damai Dai, Shuming Ma, Barun Patra, Saksham
  Singhal, Payal Bajaj, Xia Song, Xian-Ling Mao, et~al.
\newblock On the representation collapse of sparse mixture of experts.
\newblock \emph{NeurIPS}, 2022.

\bibitem[Ding et~al.(2021)Ding, Yang, Hong, Zheng, Zhou, Yin, Lin, Zou, Shao,
  Yang, et~al.]{ding2021cogview}
Ming Ding, Zhuoyi Yang, Wenyi Hong, Wendi Zheng, Chang Zhou, Da~Yin, Junyang
  Lin, Xu~Zou, Zhou Shao, Hongxia Yang, et~al.
\newblock Cogview: Mastering text-to-image generation via transformers.
\newblock \emph{NeurIPS}, 2021.

\bibitem[Don-Yehiya et~al.(2022)Don-Yehiya, Venezian, Raffel, Slonim, Katz, and
  Choshen]{don2022cold}
Shachar Don-Yehiya, Elad Venezian, Colin Raffel, Noam Slonim, Yoav Katz, and
  Leshem Choshen.
\newblock Cold fusion: Collaborative descent for distributed multitask
  finetuning.
\newblock \emph{arXiv preprint arXiv:2212.01378}, 2022.

\bibitem[Du et~al.(2022)Du, Huang, Dai, Tong, Lepikhin, Xu, Krikun, Zhou, Yu,
  Firat, et~al.]{du2022glam}
Nan Du, Yanping Huang, Andrew~M Dai, Simon Tong, Dmitry Lepikhin, Yuanzhong Xu,
  Maxim Krikun, Yanqi Zhou, Adams~Wei Yu, Orhan Firat, et~al.
\newblock Glam: Efficient scaling of language models with mixture-of-experts.
\newblock In \emph{ICML}, 2022.

\bibitem[Eigen et~al.(2013)Eigen, Ranzato, and Sutskever]{eigen2013learning}
David Eigen, Marc'Aurelio Ranzato, and Ilya Sutskever.
\newblock Learning factored representations in a deep mixture of experts.
\newblock \emph{arXiv preprint arXiv:1312.4314}, 2013.

\bibitem[Fedus et~al.(2022)Fedus, Zoph, and Shazeer]{fedus2022switch}
William Fedus, Barret Zoph, and Noam Shazeer.
\newblock Switch transformers: Scaling to trillion parameter models with simple
  and efficient sparsity.
\newblock \emph{JMLR}, 23\penalty0 (120):\penalty0 1--39, 2022.

\bibitem[Feng et~al.(2023)Feng, Zhang, Yu, Fang, Li, Chen, Lu, Liu, Yin, Feng,
  et~al.]{feng2023ernie}
Zhida Feng, Zhenyu Zhang, Xintong Yu, Yewei Fang, Lanxin Li, Xuyi Chen, Yuxiang
  Lu, Jiaxiang Liu, Weichong Yin, Shikun Feng, et~al.
\newblock Ernie-vilg 2.0: Improving text-to-image diffusion model with
  knowledge-enhanced mixture-of-denoising-experts.
\newblock In \emph{CVPR}, 2023.

\bibitem[Frankle et~al.(2020)Frankle, Dziugaite, Roy, and
  Carbin]{frankle2020linear}
Jonathan Frankle, Gintare~Karolina Dziugaite, Daniel Roy, and Michael Carbin.
\newblock Linear mode connectivity and the lottery ticket hypothesis.
\newblock In \emph{ICML}, 2020.

\bibitem[Freund \& Schapire(1995)Freund and Schapire]{freund1995desicion}
Yoav Freund and Robert~E Schapire.
\newblock A desicion-theoretic generalization of on-line learning and an
  application to boosting.
\newblock In \emph{European conference on computational learning theory}, pp.\
  23--37. Springer, 1995.

\bibitem[Fu et~al.(2024)Fu, Jiang, Liu, Wang, Deng, Chen, and Gu]{fu2024ap}
Yuchen Fu, Zhiwei Jiang, Yuliang Liu, Cong Wang, Zexuan Deng, Zhaoling Chen,
  and Qing Gu.
\newblock Ap-adapter: Improving generalization of automatic prompts on unseen
  text-to-image diffusion models.
\newblock In \emph{NeurIPS}, 2024.

\bibitem[Gafni et~al.(2022)Gafni, Polyak, Ashual, Sheynin, Parikh, and
  Taigman]{gafni2022make}
Oran Gafni, Adam Polyak, Oron Ashual, Shelly Sheynin, Devi Parikh, and Yaniv
  Taigman.
\newblock Make-a-scene: Scene-based text-to-image generation with human priors.
\newblock In \emph{ECCV}, 2022.

\bibitem[Goodfellow et~al.(2014)Goodfellow, Pouget-Abadie, Mirza, Xu,
  Warde-Farley, Ozair, Courville, and Bengio]{goodfellow2014generative}
Ian Goodfellow, Jean Pouget-Abadie, Mehdi Mirza, Bing Xu, David Warde-Farley,
  Sherjil Ozair, Aaron Courville, and Yoshua Bengio.
\newblock Generative adversarial nets.
\newblock \emph{NeurIPS}, 2014.

\bibitem[He et~al.(2016)He, Zhang, Ren, and Sun]{he2016deep}
Kaiming He, Xiangyu Zhang, Shaoqing Ren, and Jian Sun.
\newblock Deep residual learning for image recognition.
\newblock In \emph{CVPR}, 2016.

\bibitem[Hessel et~al.(2021)Hessel, Holtzman, Forbes, Le~Bras, and
  Choi]{hessel2021clipscore}
Jack Hessel, Ari Holtzman, Maxwell Forbes, Ronan Le~Bras, and Yejin Choi.
\newblock Clipscore: A reference-free evaluation metric for image captioning.
\newblock In \emph{EMNLP}, 2021.

\bibitem[Ho \& Salimans(2022)Ho and Salimans]{ho2022classifier}
Jonathan Ho and Tim Salimans.
\newblock Classifier-free diffusion guidance.
\newblock \emph{arXiv preprint arXiv:2207.12598}, 2022.

\bibitem[Ho et~al.(2020)Ho, Jain, and Abbeel]{ho2020denoising}
Jonathan Ho, Ajay Jain, and Pieter Abbeel.
\newblock Denoising diffusion probabilistic models.
\newblock \emph{NeurIPS}, 2020.

\bibitem[Ilharco et~al.(2022{\natexlab{a}})Ilharco, Ribeiro, Wortsman,
  Gururangan, Schmidt, Hajishirzi, and Farhadi]{ilharco2022editing}
Gabriel Ilharco, Marco~Tulio Ribeiro, Mitchell Wortsman, Suchin Gururangan,
  Ludwig Schmidt, Hannaneh Hajishirzi, and Ali Farhadi.
\newblock Editing models with task arithmetic.
\newblock \emph{arXiv preprint arXiv:2212.04089}, 2022{\natexlab{a}}.

\bibitem[Ilharco et~al.(2022{\natexlab{b}})Ilharco, Wortsman, Gadre, Song,
  Hajishirzi, Kornblith, Farhadi, and Schmidt]{ilharco2022patching}
Gabriel Ilharco, Mitchell Wortsman, Samir~Yitzhak Gadre, Shuran Song, Hannaneh
  Hajishirzi, Simon Kornblith, Ali Farhadi, and Ludwig Schmidt.
\newblock Patching open-vocabulary models by interpolating weights.
\newblock \emph{NeurIPS}, 2022{\natexlab{b}}.

\bibitem[Jin et~al.(2022)Jin, Ren, Preotiuc-Pietro, and Cheng]{jin2022dataless}
Xisen Jin, Xiang Ren, Daniel Preotiuc-Pietro, and Pengxiang Cheng.
\newblock Dataless knowledge fusion by merging weights of language models.
\newblock \emph{arXiv preprint arXiv:2212.09849}, 2022.

\bibitem[Kingma \& Welling(2013)Kingma and Welling]{kingma2013auto}
Diederik~P Kingma and Max Welling.
\newblock Auto-encoding variational bayes.
\newblock \emph{arXiv preprint arXiv:1312.6114}, 2013.

\bibitem[Kirstain et~al.(2023)Kirstain, Polyak, Singer, Matiana, Penna, and
  Levy]{kirstain2023pick}
Yuval Kirstain, Adam Polyak, Uriel Singer, Shahbuland Matiana, Joe Penna, and
  Omer Levy.
\newblock Pick-a-pic: An open dataset of user preferences for text-to-image
  generation.
\newblock \emph{arXiv preprint arXiv:2305.01569}, 2023.

\bibitem[Kumari et~al.(2022)Kumari, Zhang, Shechtman, and
  Zhu]{kumari2022ensembling}
Nupur Kumari, Richard Zhang, Eli Shechtman, and Jun-Yan Zhu.
\newblock Ensembling off-the-shelf models for gan training.
\newblock In \emph{CVPR}, 2022.

\bibitem[Lee et~al.(2024)Lee, Kim, Go, Jeong, Oh, and Choi]{lee2024multi}
Yunsung Lee, JinYoung Kim, Hyojun Go, Myeongho Jeong, Shinhyeok Oh, and
  Seungtaek Choi.
\newblock Multi-architecture multi-expert diffusion models.
\newblock In \emph{AAAI}, 2024.

\bibitem[Li et~al.(2022)Li, Gururangan, Dettmers, Lewis, Althoff, Smith, and
  Zettlemoyer]{li2022branch}
Margaret Li, Suchin Gururangan, Tim Dettmers, Mike Lewis, Tim Althoff, Noah~A
  Smith, and Luke Zettlemoyer.
\newblock Branch-train-merge: Embarrassingly parallel training of expert
  language models.
\newblock \emph{arXiv preprint arXiv:2208.03306}, 2022.

\bibitem[Lin et~al.(2014)Lin, Maire, Belongie, Hays, Perona, Ramanan,
  Doll{\'a}r, and Zitnick]{lin2014microsoft}
Tsung-Yi Lin, Michael Maire, Serge Belongie, James Hays, Pietro Perona, Deva
  Ramanan, Piotr Doll{\'a}r, and C~Lawrence Zitnick.
\newblock Microsoft coco: Common objects in context.
\newblock In \emph{ECCV}, 2014.

\bibitem[Liu et~al.(2022)Liu, Ren, Lin, and Zhao]{liu2022pseudo}
Luping Liu, Yi~Ren, Zhijie Lin, and Zhou Zhao.
\newblock Pseudo numerical methods for diffusion models on manifolds.
\newblock \emph{arXiv preprint arXiv:2202.09778}, 2022.

\bibitem[Loshchilov \& Hutter(2017)Loshchilov and
  Hutter]{loshchilov2017decoupled}
Ilya Loshchilov and Frank Hutter.
\newblock Decoupled weight decay regularization.
\newblock \emph{arXiv preprint arXiv:1711.05101}, 2017.

\bibitem[Lu et~al.(2022{\natexlab{a}})Lu, Zhou, Bao, Chen, Li, and
  Zhu]{lu2022dpm}
Cheng Lu, Yuhao Zhou, Fan Bao, Jianfei Chen, Chongxuan Li, and Jun Zhu.
\newblock Dpm-solver: A fast ode solver for diffusion probabilistic model
  sampling in around 10 steps.
\newblock \emph{NeurIPS}, 2022{\natexlab{a}}.

\bibitem[Lu et~al.(2022{\natexlab{b}})Lu, Zhou, Bao, Chen, Li, and
  Zhu]{lu2022dpm2}
Cheng Lu, Yuhao Zhou, Fan Bao, Jianfei Chen, Chongxuan Li, and Jun Zhu.
\newblock Dpm-solver++: Fast solver for guided sampling of diffusion
  probabilistic models.
\newblock \emph{arXiv preprint arXiv:2211.01095}, 2022{\natexlab{b}}.

\bibitem[Luo et~al.(2023)Luo, Tan, Huang, Li, and Zhao]{luo2023latent}
Simian Luo, Yiqin Tan, Longbo Huang, Jian Li, and Hang Zhao.
\newblock Latent consistency models: Synthesizing high-resolution images with
  few-step inference.
\newblock \emph{arXiv preprint arXiv:2310.04378}, 2023.

\bibitem[Matena \& Raffel(2022)Matena and Raffel]{matena2022merging}
Michael~S Matena and Colin~A Raffel.
\newblock Merging models with fisher-weighted averaging.
\newblock \emph{NeurIPS}, 2022.

\bibitem[Mendes-Moreira et~al.(2012)Mendes-Moreira, Soares, Jorge, and
  Sousa]{mendes2012ensemble}
Joao Mendes-Moreira, Carlos Soares, Al{\'\i}pio~M{\'a}rio Jorge, and Jorge
  Freire~De Sousa.
\newblock Ensemble approaches for regression: A survey.
\newblock \emph{Acm computing surveys (csur)}, 45\penalty0 (1):\penalty0 1--40,
  2012.

\bibitem[Nichol et~al.(2021)Nichol, Dhariwal, Ramesh, Shyam, Mishkin, McGrew,
  Sutskever, and Chen]{nichol2021glide}
Alex Nichol, Prafulla Dhariwal, Aditya Ramesh, Pranav Shyam, Pamela Mishkin,
  Bob McGrew, Ilya Sutskever, and Mark Chen.
\newblock Glide: Towards photorealistic image generation and editing with
  text-guided diffusion models.
\newblock \emph{arXiv preprint arXiv:2112.10741}, 2021.

\bibitem[Ortiz-Jimenez et~al.(2023)Ortiz-Jimenez, Favero, and
  Frossard]{ortiz2023task}
Guillermo Ortiz-Jimenez, Alessandro Favero, and Pascal Frossard.
\newblock Task arithmetic in the tangent space: Improved editing of pre-trained
  models.
\newblock \emph{arXiv preprint arXiv:2305.12827}, 2023.

\bibitem[Pan et~al.(2023)Pan, Sun, Ge, Li, Duan, Wu, Zhang, Zhou, Qin, Wang,
  et~al.]{pan2023journeydb}
Junting Pan, Keqiang Sun, Yuying Ge, Hao Li, Haodong Duan, Xiaoshi Wu, Renrui
  Zhang, Aojun Zhou, Zipeng Qin, Yi~Wang, et~al.
\newblock Journeydb: A benchmark for generative image understanding.
\newblock \emph{arXiv preprint arXiv:2307.00716}, 2023.

\bibitem[Peebles \& Xie(2023)Peebles and Xie]{peebles2023scalable}
William Peebles and Saining Xie.
\newblock Scalable diffusion models with transformers.
\newblock In \emph{ICCV}, 2023.

\bibitem[Podell et~al.(2023)Podell, English, Lacey, Blattmann, Dockhorn,
  M{\"u}ller, Penna, and Rombach]{podell2023sdxl}
Dustin Podell, Zion English, Kyle Lacey, Andreas Blattmann, Tim Dockhorn, Jonas
  M{\"u}ller, Joe Penna, and Robin Rombach.
\newblock Sdxl: Improving latent diffusion models for high-resolution image
  synthesis.
\newblock \emph{arXiv preprint arXiv:2307.01952}, 2023.

\bibitem[Radford et~al.(2021)Radford, Kim, Hallacy, Ramesh, Goh, Agarwal,
  Sastry, Askell, Mishkin, Clark, et~al.]{radford2021learning}
Alec Radford, Jong~Wook Kim, Chris Hallacy, Aditya Ramesh, Gabriel Goh,
  Sandhini Agarwal, Girish Sastry, Amanda Askell, Pamela Mishkin, Jack Clark,
  et~al.
\newblock Learning transferable visual models from natural language
  supervision.
\newblock In \emph{ICML}, 2021.

\bibitem[Ramesh et~al.(2021)Ramesh, Pavlov, Goh, Gray, Voss, Radford, Chen, and
  Sutskever]{ramesh2021zero}
Aditya Ramesh, Mikhail Pavlov, Gabriel Goh, Scott Gray, Chelsea Voss, Alec
  Radford, Mark Chen, and Ilya Sutskever.
\newblock Zero-shot text-to-image generation.
\newblock In \emph{ICML}, 2021.

\bibitem[Ramesh et~al.(2022)Ramesh, Dhariwal, Nichol, Chu, and
  Chen]{ramesh2022hierarchical}
Aditya Ramesh, Prafulla Dhariwal, Alex Nichol, Casey Chu, and Mark Chen.
\newblock Hierarchical text-conditional image generation with clip latents.
\newblock \emph{arXiv preprint arXiv:2204.06125}, 2022.

\bibitem[Ren et~al.(2015)Ren, Suganthan, and Srikanth]{ren2015ensemble}
Ye~Ren, PN~Suganthan, and N~Srikanth.
\newblock Ensemble methods for wind and solar power forecasting—a
  state-of-the-art review.
\newblock \emph{Renewable and Sustainable Energy Reviews}, 50:\penalty0 82--91,
  2015.

\bibitem[Rezende et~al.(2014)Rezende, Mohamed, and
  Wierstra]{rezende2014stochastic}
Danilo~Jimenez Rezende, Shakir Mohamed, and Daan Wierstra.
\newblock Stochastic backpropagation and approximate inference in deep
  generative models.
\newblock In \emph{ICML}, 2014.

\bibitem[Riquelme et~al.(2021)Riquelme, Puigcerver, Mustafa, Neumann, Jenatton,
  Susano~Pinto, Keysers, and Houlsby]{riquelme2021scaling}
Carlos Riquelme, Joan Puigcerver, Basil Mustafa, Maxim Neumann, Rodolphe
  Jenatton, Andr{\'e} Susano~Pinto, Daniel Keysers, and Neil Houlsby.
\newblock Scaling vision with sparse mixture of experts.
\newblock \emph{NeurIPS}, 2021.

\bibitem[Rokach(2010)]{rokach2010ensemble}
Lior Rokach.
\newblock Ensemble-based classifiers.
\newblock \emph{Artificial intelligence review}, 33:\penalty0 1--39, 2010.

\bibitem[Rombach et~al.(2022)Rombach, Blattmann, Lorenz, Esser, and
  Ommer]{rombach2022high}
Robin Rombach, Andreas Blattmann, Dominik Lorenz, Patrick Esser, and Bj{\"o}rn
  Ommer.
\newblock High-resolution image synthesis with latent diffusion models.
\newblock In \emph{CVPR}, 2022.

\bibitem[Ronneberger et~al.(2015)Ronneberger, Fischer, and
  Brox]{ronneberger2015u}
Olaf Ronneberger, Philipp Fischer, and Thomas Brox.
\newblock U-net: Convolutional networks for biomedical image segmentation.
\newblock In \emph{MICCAI}, 2015.

\bibitem[Saharia et~al.(2022)Saharia, Chan, Saxena, Li, Whang, Denton,
  Ghasemipour, Gontijo~Lopes, Karagol~Ayan, Salimans,
  et~al.]{saharia2022photorealistic}
Chitwan Saharia, William Chan, Saurabh Saxena, Lala Li, Jay Whang, Emily~L
  Denton, Kamyar Ghasemipour, Raphael Gontijo~Lopes, Burcu Karagol~Ayan, Tim
  Salimans, et~al.
\newblock Photorealistic text-to-image diffusion models with deep language
  understanding.
\newblock \emph{NeurIPS}, 2022.

\bibitem[Shen \& Tang(2024)Shen and Tang]{shen2024imagpose}
Fei Shen and Jinhui Tang.
\newblock Imagpose: A unified conditional framework for pose-guided person
  generation.
\newblock In \emph{NeurIPS}, 2024.

\bibitem[Shen et~al.(2023{\natexlab{a}})Shen, Ye, Zhang, Wang, Han, and
  Yang]{shen2023advancing}
Fei Shen, Hu~Ye, Jun Zhang, Cong Wang, Xiao Han, and Wei Yang.
\newblock Advancing pose-guided image synthesis with progressive conditional
  diffusion models.
\newblock \emph{arXiv preprint arXiv:2310.06313}, 2023{\natexlab{a}}.

\bibitem[Shen et~al.(2024{\natexlab{a}})Shen, Jiang, He, Ye, Wang, Du, Li, and
  Tang]{shen2024imagdressing}
Fei Shen, Xin Jiang, Xin He, Hu~Ye, Cong Wang, Xiaoyu Du, Zechao Li, and Jinhui
  Tang.
\newblock Imagdressing-v1: Customizable virtual dressing.
\newblock \emph{arXiv preprint arXiv:2407.12705}, 2024{\natexlab{a}}.

\bibitem[Shen et~al.(2024{\natexlab{b}})Shen, Ye, Liu, Zhang, Wang, Han, and
  Yang]{shen2024boosting}
Fei Shen, Hu~Ye, Sibo Liu, Jun Zhang, Cong Wang, Xiao Han, and Wei Yang.
\newblock Boosting consistency in story visualization with rich-contextual
  conditional diffusion models.
\newblock \emph{arXiv preprint arXiv:2407.02482}, 2024{\natexlab{b}}.

\bibitem[Shen et~al.(2025)Shen, Wang, Gao, Guo, Dang, Tang, and
  Chua]{shen2025long}
Fei Shen, Cong Wang, Junyao Gao, Qin Guo, Jisheng Dang, Jinhui Tang, and
  Tat-Seng Chua.
\newblock Long-term talkingface generation via motion-prior conditional
  diffusion model.
\newblock \emph{arXiv preprint arXiv:2502.09533}, 2025.

\bibitem[Shen et~al.(2023{\natexlab{b}})Shen, Yao, Li, Darrell, Keutzer, and
  He]{shen2023scaling}
Sheng Shen, Zhewei Yao, Chunyuan Li, Trevor Darrell, Kurt Keutzer, and Yuxiong
  He.
\newblock Scaling vision-language models with sparse mixture of experts.
\newblock \emph{arXiv preprint arXiv:2303.07226}, 2023{\natexlab{b}}.

\bibitem[Sohl-Dickstein et~al.(2015)Sohl-Dickstein, Weiss, Maheswaranathan, and
  Ganguli]{sohl2015deep}
Jascha Sohl-Dickstein, Eric Weiss, Niru Maheswaranathan, and Surya Ganguli.
\newblock Deep unsupervised learning using nonequilibrium thermodynamics.
\newblock In \emph{ICML}, 2015.

\bibitem[Song et~al.(2020)Song, Meng, and Ermon]{song2020denoising}
Jiaming Song, Chenlin Meng, and Stefano Ermon.
\newblock Denoising diffusion implicit models.
\newblock In \emph{ICLR}, 2020.

\bibitem[Stoica et~al.(2023)Stoica, Bolya, Bjorner, Hearn, and
  Hoffman]{stoica2023zipit}
George Stoica, Daniel Bolya, Jakob Bjorner, Taylor Hearn, and Judy Hoffman.
\newblock Zipit! merging models from different tasks without training.
\newblock \emph{arXiv preprint arXiv:2305.03053}, 2023.

\bibitem[Tao et~al.(2022)Tao, Tang, Wu, Jing, Bao, and Xu]{tao2022df}
Ming Tao, Hao Tang, Fei Wu, Xiao-Yuan Jing, Bing-Kun Bao, and Changsheng Xu.
\newblock Df-gan: A simple and effective baseline for text-to-image synthesis.
\newblock In \emph{CVPR}, 2022.

\bibitem[Vaswani et~al.(2017)Vaswani, Shazeer, Parmar, Uszkoreit, Jones, Gomez,
  Kaiser, and Polosukhin]{vaswani2017attention}
Ashish Vaswani, Noam Shazeer, Niki Parmar, Jakob Uszkoreit, Llion Jones,
  Aidan~N Gomez, {\L}ukasz Kaiser, and Illia Polosukhin.
\newblock Attention is all you need.
\newblock \emph{NeurIPS}, 2017.

\bibitem[Vega-Pons \& Ruiz-Shulcloper(2011)Vega-Pons and
  Ruiz-Shulcloper]{vega2011survey}
Sandro Vega-Pons and Jos{\'e} Ruiz-Shulcloper.
\newblock A survey of clustering ensemble algorithms.
\newblock \emph{International Journal of Pattern Recognition and Artificial
  Intelligence}, 25\penalty0 (03):\penalty0 337--372, 2011.

\bibitem[Wang et~al.(2024)Wang, Tian, Zhang, Guan, Luo, Shen, Jiang, Gu, Han,
  and Yang]{wang2024v}
Cong Wang, Kuan Tian, Jun Zhang, Yonghang Guan, Feng Luo, Fei Shen, Zhiwei
  Jiang, Qing Gu, Xiao Han, and Wei Yang.
\newblock V-express: Conditional dropout for progressive training of portrait
  video generation.
\newblock \emph{arXiv preprint arXiv:2406.02511}, 2024.

\bibitem[Wang et~al.(2022)Wang, Montoya, Munechika, Yang, Hoover, and
  Chau]{wang2022diffusiondb}
Zijie~J Wang, Evan Montoya, David Munechika, Haoyang Yang, Benjamin Hoover, and
  Duen~Horng Chau.
\newblock Diffusiondb: A large-scale prompt gallery dataset for text-to-image
  generative models.
\newblock \emph{arXiv preprint arXiv:2210.14896}, 2022.

\bibitem[Wortsman et~al.(2022{\natexlab{a}})Wortsman, Ilharco, Gadre, Roelofs,
  Gontijo-Lopes, Morcos, Namkoong, Farhadi, Carmon, Kornblith,
  et~al.]{wortsman2022model}
Mitchell Wortsman, Gabriel Ilharco, Samir~Ya Gadre, Rebecca Roelofs, Raphael
  Gontijo-Lopes, Ari~S Morcos, Hongseok Namkoong, Ali Farhadi, Yair Carmon,
  Simon Kornblith, et~al.
\newblock Model soups: averaging weights of multiple fine-tuned models improves
  accuracy without increasing inference time.
\newblock In \emph{ICML}, 2022{\natexlab{a}}.

\bibitem[Wortsman et~al.(2022{\natexlab{b}})Wortsman, Ilharco, Kim, Li,
  Kornblith, Roelofs, Lopes, Hajishirzi, Farhadi, Namkoong,
  et~al.]{wortsman2022robust}
Mitchell Wortsman, Gabriel Ilharco, Jong~Wook Kim, Mike Li, Simon Kornblith,
  Rebecca Roelofs, Raphael~Gontijo Lopes, Hannaneh Hajishirzi, Ali Farhadi,
  Hongseok Namkoong, et~al.
\newblock Robust fine-tuning of zero-shot models.
\newblock In \emph{CVPR}, 2022{\natexlab{b}}.

\bibitem[Wu et~al.(2023)Wu, Hao, Sun, Chen, Zhu, Zhao, and Li]{wu2023human}
Xiaoshi Wu, Yiming Hao, Keqiang Sun, Yixiong Chen, Feng Zhu, Rui Zhao, and
  Hongsheng Li.
\newblock Human preference score v2: A solid benchmark for evaluating human
  preferences of text-to-image synthesis.
\newblock \emph{arXiv preprint arXiv:2306.09341}, 2023.

\bibitem[Xu et~al.(2023)Xu, Liu, Wu, Tong, Li, Ding, Tang, and
  Dong]{xu2023imagereward}
Jiazheng Xu, Xiao Liu, Yuchen Wu, Yuxuan Tong, Qinkai Li, Ming Ding, Jie Tang,
  and Yuxiao Dong.
\newblock Imagereward: Learning and evaluating human preferences for
  text-to-image generation.
\newblock \emph{arXiv preprint arXiv:2304.05977}, 2023.

\bibitem[Xu et~al.(2018)Xu, Zhang, Huang, Zhang, Gan, Huang, and
  He]{xu2018attngan}
Tao Xu, Pengchuan Zhang, Qiuyuan Huang, Han Zhang, Zhe Gan, Xiaolei Huang, and
  Xiaodong He.
\newblock Attngan: Fine-grained text to image generation with attentional
  generative adversarial networks.
\newblock In \emph{CVPR}, 2018.

\bibitem[Xue et~al.(2023)Xue, Song, Guo, Liu, Zong, Liu, and
  Luo]{xue2023raphael}
Zeyue Xue, Guanglu Song, Qiushan Guo, Boxiao Liu, Zhuofan Zong, Yu~Liu, and
  Ping Luo.
\newblock Raphael: Text-to-image generation via large mixture of diffusion
  paths.
\newblock \emph{arXiv preprint arXiv:2305.18295}, 2023.

\bibitem[Yang et~al.(2010)Yang, Hwa~Yang, B~Zhou, and Y~Zomaya]{yang2010review}
Pengyi Yang, Yee Hwa~Yang, Bing B~Zhou, and Albert Y~Zomaya.
\newblock A review of ensemble methods in bioinformatics.
\newblock \emph{Current Bioinformatics}, 5\penalty0 (4):\penalty0 296--308,
  2010.

\bibitem[Yu et~al.(2022)Yu, Xu, Koh, Luong, Baid, Wang, Vasudevan, Ku, Yang,
  Ayan, et~al.]{yu2022scaling}
Jiahui Yu, Yuanzhong Xu, Jing~Yu Koh, Thang Luong, Gunjan Baid, Zirui Wang,
  Vijay Vasudevan, Alexander Ku, Yinfei Yang, Burcu~Karagol Ayan, et~al.
\newblock Scaling autoregressive models for content-rich text-to-image
  generation.
\newblock \emph{arXiv preprint arXiv:2206.10789}, 2022.

\bibitem[Zhang et~al.(2021{\natexlab{a}})Zhang, Koh, Baldridge, Lee, and
  Yang]{zhang2021cross}
Han Zhang, Jing~Yu Koh, Jason Baldridge, Honglak Lee, and Yinfei Yang.
\newblock Cross-modal contrastive learning for text-to-image generation.
\newblock In \emph{CVPR}, 2021{\natexlab{a}}.

\bibitem[Zhang et~al.(2021{\natexlab{b}})Zhang, Yin, Fang, Li, Duan, Wu, Sun,
  Tian, Wu, and Wang]{zhang2021ernie}
Han Zhang, Weichong Yin, Yewei Fang, Lanxin Li, Boqiang Duan, Zhihua Wu,
  Yu~Sun, Hao Tian, Hua Wu, and Haifeng Wang.
\newblock Ernie-vilg: Unified generative pre-training for bidirectional
  vision-language generation.
\newblock \emph{arXiv preprint arXiv:2112.15283}, 2021{\natexlab{b}}.

\bibitem[Zhang et~al.(2023)Zhang, Chen, Liu, and He]{zhang2023composing}
Jinghan Zhang, Shiqi Chen, Junteng Liu, and Junxian He.
\newblock Composing parameter-efficient modules with arithmetic operations.
\newblock \emph{arXiv preprint arXiv:2306.14870}, 2023.

\bibitem[Zhao et~al.(2023)Zhao, Zheng, Wang, Lan, and
  Yang]{zhao2023magicfusion}
Jing Zhao, Heliang Zheng, Chaoyue Wang, Long Lan, and Wenjing Yang.
\newblock Magicfusion: Boosting text-to-image generation performance by fusing
  diffusion models.
\newblock \emph{arXiv preprint arXiv:2303.13126}, 2023.

\bibitem[Zhao et~al.(2005)Zhao, Gao, and Yang]{zhao2005survey}
Ying Zhao, Jun Gao, and Xuezhi Yang.
\newblock A survey of neural network ensembles.
\newblock In \emph{2005 international conference on neural networks and brain},
  volume~1, pp.\  438--442. IEEE, 2005.

\bibitem[Zhou(2012)]{zhou2012ensemble}
Zhi-Hua Zhou.
\newblock \emph{Ensemble methods: foundations and algorithms}.
\newblock CRC press, 2012.

\bibitem[Zhu et~al.(2019)Zhu, Pan, Chen, and Yang]{zhu2019dm}
Minfeng Zhu, Pingbo Pan, Wei Chen, and Yi~Yang.
\newblock Dm-gan: Dynamic memory generative adversarial networks for
  text-to-image synthesis.
\newblock In \emph{CVPR}, 2019.

\end{thebibliography}
\bibliographystyle{iclr2025_conference}

\appendix

\section{Algorithm of AFA}
\label{app:algo}

The forward process of AFA can be seen in Alg.~\ref{alg}.

\begin{algorithm}[ht]
    \caption{One forward process of Adaptive Feature Aggregation (AFA).}
    \label{alg}
    \begin{algorithmic}[1]
        \STATE{\textbf{Input:} 
            Timestep $t$,
            Condition $\mathbf{c}$,
            Input latent $\mathbf{x}_t$,
            U-Net denoisers $\{\boldsymbol{\epsilon}_{\theta_i}\}_{i=1}^N$, each of which contains $K$ blocks, and
            SABW $f_\varphi^{(j)}$ for $j$-th U-Net block.
        }
        \STATE{\textbf{Output:} 
            Predicted noise $\hat{\boldsymbol{\epsilon}}_t$.
        }
        \STATE{Initialize an empty stack $\mathcal{S}$ to store the features from the skip connections;}
        \STATE{$\mathbf{x}_{t}^{(1)}\leftarrow \mathbf{x}_t$;}
        \FOR{block index $j$ from 1 to $K$}
            {\color[rgb]{0.04, 0.68, 0.93}
            \STATE{$\vartriangleright$ \textit{Ensembling the $j$-th U-Net block by SABW feature aggregator $f_\varphi^{(j)}$}}
            \FOR{U-Net denoiser index $i$ from 1 to $N$}
                \STATE{Get the output feature $\mathbf{y}_{t,i}^{(j)}$ base on Eq.~\ref{eq:unet-denoiser};}
            \ENDFOR
            \STATE{Get $\mathbf{y}_t^{(j)}$ by aggregating all $\mathbf{y}_{t,i}^{(j)}$ base on Eq.~\ref{eq:ba-adpter} for the next block;}}
            \IF{$(j+1)$-th block is down-sampling block}
                \STATE{$\mathbf{x}_{t}^{(j+1)}\leftarrow \mathbf{y}_t^{(j)}$ and push $\mathbf{y}_t^{(j)}$ into $\mathcal{S}$;}
            \ELSIF{$(j+1)$-th block is up-sampling block}
                \STATE{Pop $\mathbf{y}_t^{s}$ from $\mathcal{S}$ and $\mathbf{x}_{t}^{(j+1)}\leftarrow \text{concat}(\mathbf{y}_t^{(j)}, \mathbf{y}_t^{s})$;}
            \ELSE
                \STATE{$\mathbf{x}_{t}^{(j+1)}\leftarrow \mathbf{y}_t^{(j)}$;}
            \ENDIF
        \ENDFOR
        \STATE{$\hat{\boldsymbol{\epsilon}}_t\leftarrow\mathbf{x}_{t}^{(K)}$;}
        \RETURN{predicted noise $\hat{\boldsymbol{\epsilon}}_t$.}
    \end{algorithmic}
\end{algorithm}

\section{Details about Parameters}
\label{app:details-about-parameters}

The parameters of AFA are concentrated in the SABW module.
The SABW module comprises three components, which are a ResNet layer, a Transformer layer, and a convolution layer, respectively.
Among them, the number of parameters in both the ResNet layer and the convolution layer varies depending on the number of ensembled models.
For instance, when AFA ensembles two diffusion models, SABW contains approximately 42.425 million parameters.
With each additional diffusion model ensembled, the parameter count of SABW increases by about 2.707 million.
Therefore, when three diffusion models are ensembled, the total number of the trainable parameters is 45.132 million, and with four models, it rises to approximately 47.839 million.
In conclusion, we estimate that AFA encompasses close to 50 million trainable parameters.

\section{Details about Evaluation Protocols}
\label{app:details-about-evaluation-protocols}

\textbf{COCO 2017} comprises 118,287 and 5,000 image-caption pairs in the test and validation sets. 
All the models generate images with a resolution of 256$\times$256.
We apply four metrics to evaluate the generation performance, which are Fréchet Inception Distance (FID), Inception Score (IS), CLIP-I, and CLIP-T, respectively.
FID and IS are applied to the test set, while CLIP-I and CLIP-T are applied to the validation set.
Both FID and IS assess the quality of generated images.
Note that lower FID means better quality.
CLIP-I is the similarity between the CLIP \citep{radford2021learning} images embeddings of generated images and that of images from the image-caption pairs.
CLIP-T is the CLIPScore \citep{hessel2021clipscore} between the generated images with the captions.

\textbf{Draw Bench Prompts} contains 200 evaluation prompts. 
All the models generate images with a resolution of 512$\times$512.
We apply 4 evaluation metrics, which are AES, Pick Score (PS) \citep{kirstain2023pick}, HPSv2 \citep{wu2023human}, and Image Reward (IR) \citep{xu2023imagereward}, respectively. 
All three metrics evaluate the performance by a model that simulates human preferences.
The evaluations are conducted 20 times to ensure statistical significance.
And the averaged metrics are reported.

\section{Effect of More Training Samples}
\label{app:effect-of-more-training-samples}

\begin{figure}[ht]
    \centering
    \includegraphics[width=0.6\columnwidth]{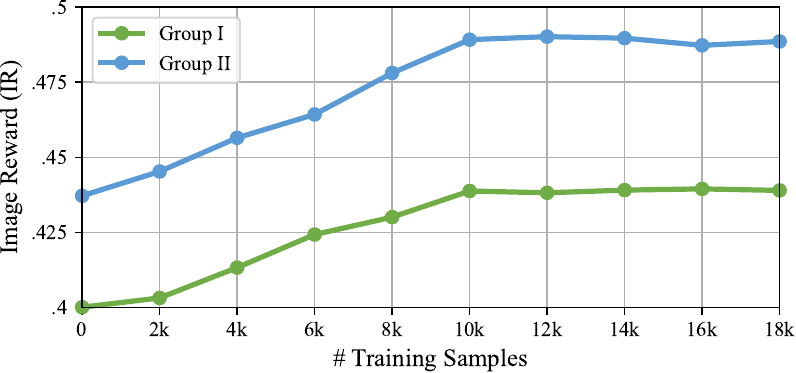}
    \vspace{-5pt}
    \caption{Effect of varying training samples.}
    \vspace{-5pt}
    \label{fig:effect-num-data}
\end{figure}

We conduct an experiment to assess the influence of an increased number of training samples on the performance of AFA. 
For this purpose, we vary the number of training samples from 0 to 18,000, increasing the increment of 2,000.
As depicted in Figure~\ref{fig:effect-num-data}, the trend of the lines initially rises and then stabilizes after reaching 10,000 training samples. 
This suggests that approximately 10,000 training samples are sufficient for AFA.
More training samples do not bring greater performance.
It may be due to the inherent limitations in the generative capabilities of the base models.

\section{More Qualitative Comparisons}

\begin{figure}[ht]
    \centering
    \includegraphics[width=1\columnwidth]{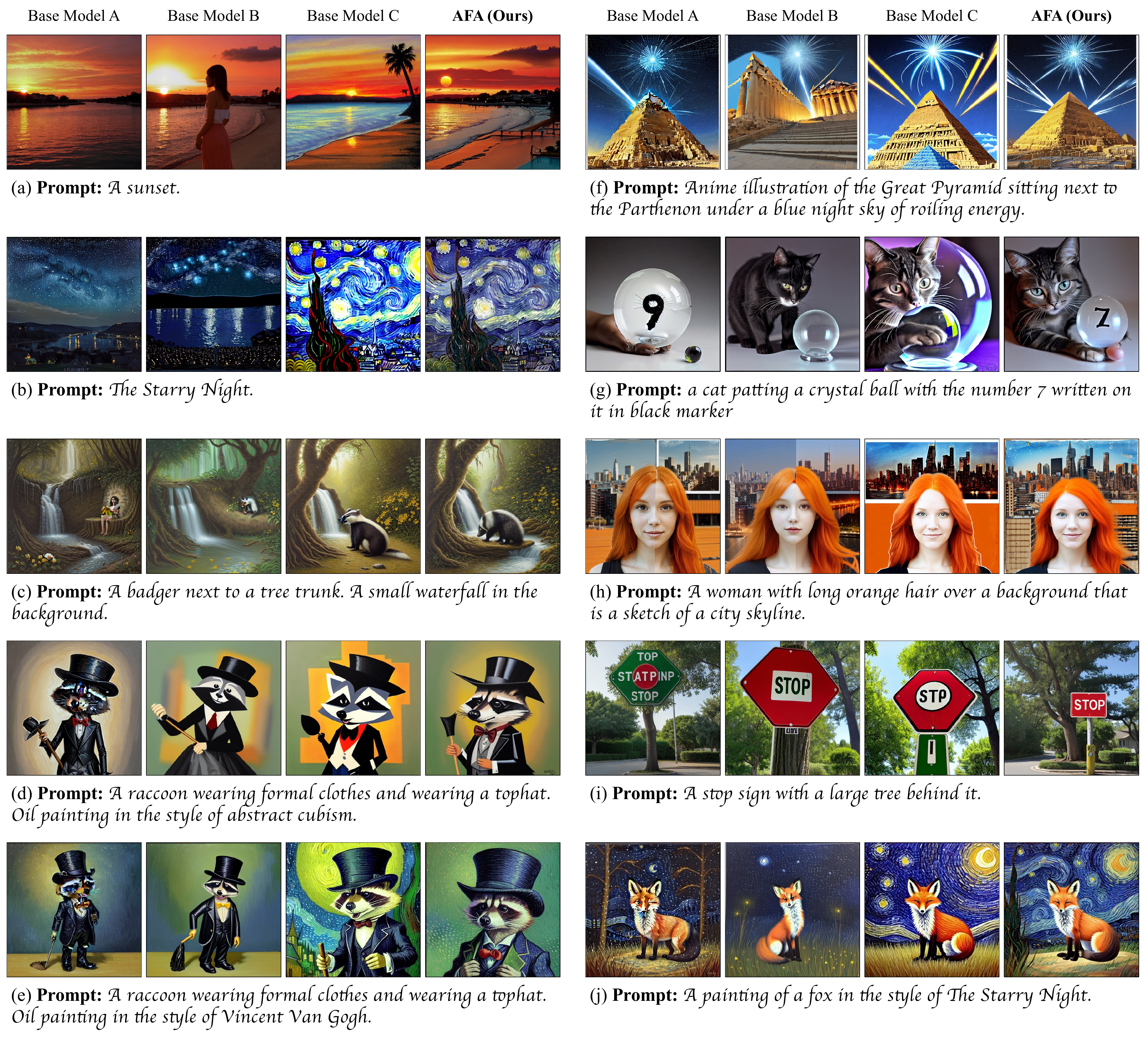}
    \vspace{-5pt}
    \caption{More qualitative comparisons between our AFA and the base models.}
    \vspace{-5pt}
    \label{fig:more-qualitative-comparisons}
\end{figure}

More qualitative comparisons between our AFA and the base models can be found in Figure~\ref{fig:more-qualitative-comparisons}.

\section{Ensembling by Mixture-of-Experts}

Another intuitive approach to ensembling multiple models involves the use of the Mixture-of-Experts (MoE) method \cite{shen2023scaling, eigen2013learning, riquelme2021scaling, fedus2022switch, du2022glam, chi2022representation}. 
The MoE framework incorporates multiple models, each designated as an \textit{expert}. 
Given an input, MoE employs a routing mechanism to determine the most suitable expert.
The input is then processed by the selected expert, and its output is considered as the output of the entire model.
The number of parameters in MoE proportionally increases with the number of ensembled experts. 
However, the parameters of MoE are sparse, indicating that not all parameters are utilized during the inference process.
Despite ensembling multiple models, only one of them is activated during inference, which makes MoE efficient.

Several text-to-image generation methods leverage MoE-based diffusion models, where multiple denoisers are ensembled along the timesteps.
In these methods, only one denoising expert is activated at each timestep.
For instance, ERNIE-ViLG 2.0 \citep{feng2023ernie} and eDiff-I \citep{balaji2022ediffi} divide all the timesteps into blocks, each of which consists of consecutive timesteps and is assigned to a denoising expert.
Building on this idea, MEME \citep{lee2024multi} introduces denoisers with different architectures, tailored to distinct timestep intervals.
In these MoE-based methods, the denoising capability of each denoiser is restricted into some specific consecutive timesteps through large-scale full-parameters training, which is not suitable to the task of model ensembling.

Inspired by the MoE-based diffusion models discussed above, an intuitive MoE-based ensembling method is to employ a trained router to select a base denoiser for the current timesteps.
We refer to this as the \textit{Denoiser-Level MoE} method.
Furthermore, to mirror the concept of our AFA, we propose another strategy of \textit{Block-Level MoE} method, which uses a router to select the next block.
The key difference between the Block-Level MoE method and our AFA lies in their operations: AFA manipulates the outputs of all blocks, whereas MoE directly acts on the input.
We will introduce the details of Block-Level MoE method, as the Denoiser-Level MoE method can be considered a special case of \textit{Block-Level MoE} method, where the selection occurs only before the first block.

Given $N$ diffusion models, we employ a router $r_\zeta^{(j)}$ to determine which model's block will be used, where $\zeta$ denotes the learnable parameters.
Specifically, $r_\zeta^{(j)}$ output the logits of $N$ models,
\begin{equation}
    \mathbf{l}_t^{(j)} = r_\zeta^{(j)}(\mathbf{x}_t^{(j)}, \mathbf{c}, t)\in\mathbb{R}^{N} \ .
\end{equation}

During training, the probability for the $i$-th model is defined by the Gumbel softmax,
\begin{equation}
    p_i = \frac{\exp((l_{t,i}^{(j)}+g)/\tau)}{\sum_{k=1}^N \exp((l_{t,k}^{(j)}+g)/\tau)} \ ,
\end{equation}
where, $l_{t,i}^{(j)}$ is the $i$-th element of $\mathbf{l}_t^{(j)}$.
$g$ is the Gumbel noise.
$\tau$ is the temperature coefficient.
To address the issue of back-propagation incapability, we apply the reparameterization trick.
Given the probability distribution $\mathbf{p}=[p_i]_{i=1}^N$, the reparameterized distribution is 
\begin{equation}
    \mathbf{p}^\prime = \text{onehot}(\mathbf{p})+\mathbf{p}-\text{sg}(\mathbf{p}) \ .
\end{equation}
Among them, $\text{onehot}(\cdot)$ is the one-hot function, which sets the maximum probability to 1 and the other to 0.
$\text{sg}(\cdot)$ is the stop gradient function.
During the forward process, the distribution is defined as $\mathbf{p}^\prime = \text{onehot}(\mathbf{p})$.
However, during the backward process, the distribution of $\mathbf{p}^\prime = \mathbf{p}$ is used to compute the gradients.
The output of the $j$-th block is 
\begin{equation}
    \label{eq:moe-output}
    \mathbf{y}_t^{(j)} = \sum\mathbf{p}^\prime \otimes \text{concat}(\{\mathbf{y}_{t,i}^{(j)}\}_{i=1}^N) \ .
\end{equation}
From Eq.~\ref{eq:moe-output}, we can infer that the forward time is proportional to the number of ensembled models during training.
The training loss is the denoising loss.

During inference, only the model that achieves maximum logits will be used to process the block input,
\begin{equation}
    \mathbf{y}_t^{(j)} = \boldsymbol{\epsilon}^{(j)}_{\theta_k}(\mathbf{x}_t^{(j)}, \mathbf{c}, t) \ , \ \ \ \ 
    k = \arg\max_{i\in[N]} \ l_{t,i}^{(j)} \ .
\end{equation}
The forward time of the ensembled model is the same as that of a single model during inference.

To implement the router $r_\zeta^{(j)}$, we apply a ResNet layer to introduce timesteps by adding the time embedding into the feature,
and a Transformer layer to introduce textual prompts by the cross-attention.
Specifically, given the input for the $j$-th block (i.e., $\mathbf{x}_t^{(j)}\in\mathbb{R}^{h_j\times w_j \times c_j}$), the output logits for $N$ models (i.e,, $\mathbf{l}_t^{(j)}$) are achieved by 
\begin{align}
    \mathbf{h}_t^{(j)} &\leftarrow \text{ResLayer}(\mathbf{x}_t^{(j)}+\gamma(t)) \in \mathbb{R}^{h_j\times w_j\times d} \ , \\
    \mathbf{o}_t^{(j)} &\leftarrow \text{TransformerLayer}(\mathbf{h}_t^{(j)},\mathbf{c})\in\mathbb{R}^{h_j\times w_j\times d} \ , \\
    \mathbf{l}_t^{(j)} &\leftarrow \text{Linear}(\text{AvgPool}(\mathbf{o}_t^{(j)}))\in\mathbb{R}^{N} \ .
\end{align}

Although the time required for a single forward pass in the MoE methods is approximately equivalent to that of a single model, they face challenges in performing parallel computations when generating multiple images simultaneously. 
This limitation becomes evident during classifier-free guidance. 
If the router selects two different models to process the conditioned and unconditioned inputs, MoE can only execute two sequential forward passes, rather than concurrently processing these two scenarios within a single forward pass.
Moreover, when generating $K$ images at once, the MoE method has a high probability of necessitating $2K$ forward passes.
While these $2K$ inputs can be divided among $N$ models, this only reduces the complexity associated with the number of forward passes from $O(K)$ to $O(N)$.
This reduction still represents a substantial computational demand.

The quantitative comparison among the base models, the MoE methods, and our AFA is summarized in Table~\ref{tab:compare-with-moe}.
AFA demonstrates superior performance compared to the MoE methods.
This may be because the base denoisers have well-balanced denoising capabilities across all timesteps, and selecting a specific denoiser for a given timestep does not significantly enhance the overall denoising capability of the ensembled model.
Furthermore, we compare AFA and the MoE methods using fewer inference steps (i.e., 20 steps).
As shown in Table~\ref{tab:compare-with-moe-20}, unlike our AFA, the MoE methods do not demonstrate robustness with fewer inference steps, performing similarity to the base models.

The efficiency comparison among the base models, the MoE methods, and our AFA is summarized in Table~\ref{tab:compare-with-moe-efficiency}.
Firstly, the MoE methods have a similar number of parameters and training TFLOPs as our AFA. 
This is because the MoE methods still require all base model to be available for selection, and during training, each base model must perform a forward pass.
Secondly, during inference, the MoE methods exhibit lower TFLOPs compared to AFA, as only one base model is activated at each timestep.
However, despite this advantage in TFLOPs, the inference time of the MoE methods is comparable to that of AFA due to their inability to leverage parallel inference effectively.
Moreover, the MoE methods do not support fewer inference steps. 
In contrast, when AFA reduces the number of inference steps to 20, it still outperforms the MoE methods, even when they use 50 inference steps, while also achieving lower TFLOPs and faster inference. 
This highlights the efficiency and effectiveness of our AFA.

\begin{table}[ht] \setlength{\tabcolsep}{9.85pt}
  \centering
  {\scriptsize
  \begin{tabular}{lcccccccc}
    \toprule
     & \multicolumn{4}{c}{Group I} & \multicolumn{4}{c}{Group II} \\ \cmidrule(lr){2-5} \cmidrule(lr){6-9}
     & AES & PS & HPSv2 & IR & AES & PS & HPSv2 & IR \\ \midrule
    Base Model A & 
    5.4102 & 21.6279 & 27.8007 & .3544 &
    5.5641 & 21.8013 & 28.0183 & .4238 \\
    Base Model B & 
    5.5013 & 21.4624 & 27.7246 & .2835 &
    5.5027 & 21.7249 & 28.0343 & .4202 \\
    Base Model C & 
    5.4881 & 21.8031 & 27.9652 & .3922 & 
    5.5712 & 21.4936 & 27.8089 & .3367 \\ \midrule \midrule
    Denoiser-Level MoE & 
    {5.4727} & {21.6893} & {27.5345} & {.4019} &
    {5.5638} & {21.5531} & {27.7843} & {.4380} \\
    {Block-Level} MoE & 
    5.5037 & 21.7690 & 27.8749 & .4037 &
    5.5718 & 21.7636 & 27.9842 & .4463 \\ \midrule
    \textbf{AFA (Ours)} &
    \textbf{5.5201} & \textbf{21.8263} & \textbf{27.9734} & \textbf{.4388} &
    \textbf{5.5798} & \textbf{21.8059} & \textbf{28.0371} & \textbf{.4892} \\
    \bottomrule
  \end{tabular}}
  \caption{Quantitative comparison between AFA with the base models and the MoE methods.}
  \label{tab:compare-with-moe}
\end{table}

\begin{table}[ht] \setlength{\tabcolsep}{9.85pt}
  \centering
  {\scriptsize
  \begin{tabular}{lcccccccc}
    \toprule
    & \multicolumn{4}{c}{Group I} & \multicolumn{4}{c}{Group II} \\ \cmidrule(lr){2-5} \cmidrule(lr){6-9}
    & AES & PS & HPSv2 & IR & AES & PS & HPSv2 & IR \\ \midrule
    Denoiser-Level MoE & 
    1.2310 & 2.9435 & 4.1739 & .0184 & 2.4556 & 3.6723 & 5.3415 & .0237 \\
    Block-Level MoE & 
    1.6854 & 3.4573 & 5.3583 & .0098 & 2.5734 & 4.2368 & 4.7252 & .0344 \\ \midrule
    \textbf{AFA (Ours)} &
    \textbf{5.4951} & \textbf{21.8093} & \textbf{27.7347} & \textbf{.4191} &
    \textbf{5.5322} & \textbf{21.7830} & \textbf{27.9775} & \textbf{.4803} \\
    \bottomrule
  \end{tabular}}
  \caption{{Quantitative comparison of AFA and the MoE methods with 20 inference steps.}}
  \label{tab:compare-with-moe-20}
\end{table}

\begin{table}[ht] \setlength{\tabcolsep}{5.45pt}
  \centering
  {\scriptsize
  \begin{tabular}{lcccccccc}
    \toprule
     & & & \multicolumn{2}{c}{1 image} & \multicolumn{2}{c}{2 images} & \multicolumn{2}{c}{4 images} \\ \cmidrule(lr){4-5} \cmidrule(lr){6-7} \cmidrule(lr){8-9}
     & \# params.$^\dagger$ & T-TLOPs & I-TFLOPs & Times (s)$^\ddagger$ & I-TFLOPs & Times (s)$^\ddagger$ & I-TFLOPs & Times (s)$^\ddagger$ \\ \midrule
    Base Model$^*$ & 
    859.52M  & -- &  70.24 & 2.94 & 140.45 &  5.61 & 280.88 &  6.87 \\ \midrule 
    Denoiser-Level MoE$^*$ & 
    2581.01M & 6.36 & 72.37 & 5.36 & 144.75 & 10.37 & 289.48 & 20.81 \\ 
    Block-Level MoE$^*$ & 
    2621.03M & 6.43 & 74.41 & 5.39 & 148.82 & 10.43 & 297.65 & 21.00 \\ \midrule
    AFA (50 inf. steps) &
    2621.03M & 6.42 & 218.74 & 8.98 & 437.48 & 15.57 & 874.96 & 21.33 \\
    AFA (20 inf. steps) &
    2621.03M & 6.42 & 86.61 & 3.62 & 173.22 &  6.41 & 346.45 &  8.74 \\
    \bottomrule
  \end{tabular}}
  \caption{Efficiency Comparison between AFA with the base models and the MoE methods when ensembling three base models to generate images with a resolution of 512$\times$512. T-TFLOPs and I-TFLOPs denote TFLOPs for training and inference, respectively. $^\dagger$Note that only the parameters of denoisers are considered. $^\ddagger$Note that all the inference times are evaluated in our A6000 environment over 20 runs. $^*$Note that both the base models and MoE methods are evaluated with 50 inference steps, because they cannot achieve comparable performance with fewer inference steps.}
  \label{tab:compare-with-moe-efficiency}
\end{table}

\section{Ensembling More Models}

To evaluate the scalability of our AFA, we apply it to ensemble a larger number of base models.
Specifically, we select and ensemble several base models with best performance, and test IR metrics in Draw Bench Prompts.
As shown in Table~\ref{tab:comparison-base-models-and-inference-steps}, ensembling more base models leads more performance improvements. Additionally, increasing the number of the ensembled base models enhances the tolerance for fewer inference steps.

\begin{table}[ht] \setlength{\tabcolsep}{8pt}
  \centering
  {\scriptsize
  \begin{tabular}{lcccccc}
    \toprule
    & 1 base model & 2 base models & 3 base models & 4 base models & 5 base models & 6 base models \\ \midrule
    50 inference steps &  
    \cellcolor{gray}{0.4238} & \cellcolor{gray}{0.5003} & \cellcolor{gray}{0.4892} & \cellcolor{gray}{0.4967} & \cellcolor{gray}{0.5042} & \cellcolor{gray}{0.5347} \\
    40 inference steps &  
    0.4065 & \cellcolor{gray}{0.4834} & \cellcolor{gray}{0.4874} & \cellcolor{gray}{0.4906} & \cellcolor{gray}{0.4984} & \cellcolor{gray}{0.5310} \\
    30 inference steps &  
    0.3593 & \cellcolor{gray}{0.4791} & \cellcolor{gray}{0.4846} & \cellcolor{gray}{0.4853} & \cellcolor{gray}{0.4993} & \cellcolor{gray}{0.5294} \\
    20 inference steps &  
    0.2103 & 0.3641 & \cellcolor{gray}{0.4803} & \cellcolor{gray}{0.4840} & \cellcolor{gray}{0.4975} & \cellcolor{gray}{0.5304} \\
    10 inference steps & 
    -0.5255 & 0.1844 & 0.4013 & 0.4285 & \cellcolor{gray}{0.4423} & \cellcolor{gray}{0.5135} \\
    \bottomrule
  \end{tabular}}
  \caption{{Performance (IR) comparison across different numbers of ensembled base models and varying inference steps.}}
  \label{tab:comparison-base-models-and-inference-steps}
\end{table}

\section{{Comparison on Fewer Inference Steps}}

We compare our AFA with the baseline methods under fewer inference steps.
As shown in Table~\ref{tab:compare-fewer-inference-steps}, our AFA not only outperforms the baseline methods at higher inference steps but also demonstrates significantly better performance under fewer inference steps. 
This highlights AFA's superior tolerance to the fewer inference steps compared to the baselines.

\begin{table}[ht] \setlength{\tabcolsep}{7pt}
  \centering
  {\scriptsize
  \begin{tabular}{lcccccccccc}
    \toprule
    & \multicolumn{5}{c}{Group I} & \multicolumn{5}{c}{Group II} \\ \cmidrule(lr){2-6} \cmidrule(lr){7-11}
    inference steps & 50 & 40 & 30 & 20 & 10 & 50 & 40 & 30 & 20 & 10 \\ \midrule
    Wtd. Merging & 
    0.3909 & 0.3104 & 0.2451 & -0.0535 & -0.7536 & 0.4387 & 0.4123 & 0.2942 &  0.1046 & -0.5105 \\ 
    MBW & 
    0.3922 & 0.3041 & 0.2345 &  0.0341 & -0.7731 & 0.4396 & 0.4094 & 0.2893 & -0.0031 & -0.5524 \\
    autoMBW &
    0.3672 & 0.3158 & 0.2510 & -0.4173 & -0.8046 & 0.3513 & 0.3545 & 0.2502 & -0.0841 & -0.5841 \\ \midrule 
    MagicFusion & 
    0.3317 & 0.3245 & 0.3104 &  0.2349 &  0.0175 & 0.4194 & 0.3995 & 0.3408 & 0.1951 & 0.0818 \\
    AFA (Ours) & 
    \textbf{0.4388} & \textbf{0.4389} & \textbf{0.4238} & \textbf{0.4191} & \textbf{0.3575} & \textbf{0.4892} & \textbf{0.4874} & \textbf{0.4846} & \textbf{0.4803} & \textbf{0.4013} \\
    \bottomrule
  \end{tabular}}
  \caption{{Quantitative comparison (IR) on fewer inference steps.}}
  \label{tab:compare-fewer-inference-steps}
\end{table}

\section{{Efficiency Comparison}}

We compare the efficiency of merging methods and ensembling methods.
As shown in Table~\ref{tab:compare-with-baselines-efficiency}, when using the same 50 inference steps, ensembling methods (e.g., MagicFusion and AFA) do not have an efficiency advantage over a single base models or merging methods. The number of model parameters, TFLOPs, and inference time all increase linearly with the number of base models.

However, thanks to AFA’s high tolerance for fewer inference steps, it can achieve similar performance with reduced steps, as demonstrated in Table~\ref{tab:comparison-base-models-and-inference-steps}, while the baseline methods fail to maintain performance, as shown in Table~\ref{tab:compare-fewer-inference-steps}.
Therefore, under fewer inference steps, AFA achieves comparable efficiency (TFLOPs and inference time) to that of a single base models and merging methods.

\begin{table}[ht] \setlength{\tabcolsep}{4.3pt}
  \centering
  {\scriptsize
  \begin{tabular}{lccccccccc}
    \toprule
    & \multicolumn{3}{c}{2 base models} & \multicolumn{3}{c}{3 base models} & \multicolumn{3}{c}{6 base models} \\ \cmidrule(lr){2-4} \cmidrule(lr){5-7} \cmidrule(lr){8-10}
    & \# params.$^\dagger$ & TFLOPs & Times (s)$^\ddagger$ & \# params.$^\dagger$ & TFLOPs & Times (s)$^\ddagger$ & \# params.$^\dagger$ & TFLOPs & Times (s)$^\ddagger$ \\ \midrule
    Merging Method$^*$ & 
     859.52M &  70.24 & 2.94 &  859.52M &  70.24 & 2.94 &  859.52M &  70.24 &  2.94 \\ 
    MagicFusion$^*$ & 
    1719.04M & 138.66 & 6.01 & 2578.56M & 206.88 & 9.13 & 5157.12M & 411.53 & 22.41 \\ \midrule 
    AFA (50 inf. steps) &
    1761.47M & 148.64 & 6.28 & 2621.03M & 218.74 & 8.98 & 5210.37M & 430.08 & 23.15 \\
    AFA (30 inf. steps) &
    1761.47M &  94.24 & 3.21 &       -- &     -- &   -- &       -- &     -- &    -- \\
    AFA (20 inf. steps) &
    --       &    -- &    -- & 2621.03M &  86.61 & 3.62 &       -- &     -- &    -- \\
    AFA (10 inf. steps) &
    --       &    -- &    -- &       -- &     -- &   -- & 5210.37M & 103.68 &  4.58 \\
    \bottomrule
  \end{tabular}}
  \caption{{Efficiency Comparison between the merging methods and the ensembling methods when generating an image with a resolution of 512$\times$512. $^\dagger$Note that only the parameters of denoisers are considered. $^\ddagger$Note that all the inference times are evaluated in our A6000 environment over 20 runs. $^*$Note that both the merging methods and MagicFusion are evaluated with 50 inference steps, because they cannot achieve comparable performance with fewer inference steps.}}
  \label{tab:compare-with-baselines-efficiency}
\end{table}

\section{{Generality on Other Architectures}}

To evaluate the generality of our AFA on other diffusion architectures, we select three SDXL \citep{podell2023sdxl} models and two FLUX .1 [dev]\footnote{\url{https://blackforestlabs.ai/announcing-black-forest-labs/}} models from CivitAI.

SDXL consists two U-Nets: the denoising U-Net and the refining U-Net.
Both U-Nets shares a similar architecture of SDv1.5, but contain larger blocks.
Consistent with the method used for ensembling SDv1.5 models, we employ the SABW module to aggregate features from each block.

FLUX .1 [dev] is a DiT-based model \citep{peebles2023scalable}.
To ensemble models with this architecture, we also utilize the SABW module to aggregate features from each DiT Transformer block.

The selected SDXL models are NoobAI-XL\footnote{\url{https://civitai.com/models/833294?modelVersionId=1070239}}, iNiverse-Mix\footnote{\url{https://civitai.com/models/226533?modelVersionId=608842}}, and epiCRealism-XL\footnote{\url{https://civitai.com/models/277058?modelVersionId=1074830}}.
The selected FLUX .1 [dev] models are PixelWave\footnote{\url{https://civitai.com/models/141592?modelVersionId=992642}} and VerusVision\footnote{\url{https://civitai.com/models/883426?modelVersionId=988886}}.

As shown in Table~\ref{tab:compare-on-other-architectures}, when ensembling models with the SDXL or FLUX .1 [dev] architectures, the performance of the ensembled model surpasses that of the individual base models, demonstrating the generality of our AFA on different model architectures. 
However, the observed improvement is relatively small, likely because the base models already exhibit strong performance, and ensembling leads to only marginal gains. 
Additionally, due to the large size of the base models, ensembling multiple such models may be less practical.

It is worth noting that our AFA cannot be applied to ensembling models with different architectures, as the misalignment of block features makes block-wise aggregation challenging. 
We view ensembling models with diverse architectures as a promising direction for future research.


\begin{table}[ht] \setlength{\tabcolsep}{10.8pt}
  \centering
  {\scriptsize
  \begin{tabular}{lcccccccc}
    \toprule
    & \multicolumn{4}{c}{SDXL} & \multicolumn{4}{c}{FLUX .1 [dev]} \\ \cmidrule(lr){2-5} \cmidrule(lr){6-9}
    & AES & PS & HPSv2 & IR & AES & PS & HPSv2 & IR \\ \midrule
    Base Model A & 
    6.1948 & 21.7854 & 28.0452 & .6314 &
    6.2341 & 21.8843 & 28.0958 & .7341 \\
    Base Model B & 
    6.1363 & 21.7783 & 28.0531 & .6328 & 
    6.2346 & 21.8593 & 28.1003 & .7324 \\
    Base Model C & 
    6.1852 & 21.7752 & 27.9984 & .6339 &
    --     & --      & --      & --    \\ \midrule
    \textbf{AFA (Ours)} &
    \textbf{6.2190} & \textbf{21.7959} & \textbf{28.0894} & \textbf{.6342} &
    \textbf{6.2490} & \textbf{21.8931} & \textbf{28.1194} & \textbf{.7370} \\
    \bottomrule
  \end{tabular}}
  \caption{{Quantitative comparison of AFA with different diffusion architectures.}}
  \label{tab:compare-on-other-architectures}
\end{table}

\section{{Distillation to Fewer Inference Steps}}

While our AFA demonstrates robustness to fewer inference steps, we aim to further distill the ensembled model into significantly fewer steps. 
Specifically, we explore using LCM \citep{luo2023latent} to achieve this.
In our experiments, we kept the parameters of the denoisers frozen while training only the parameters of the SABW modules. 
However, as shown in Table~\ref{tab:reuslts-distill}, this approach is unsuccessful. 
A possible reason for this failure is that the frozen parameters may have impeded the distillation process.

\begin{table}[ht] \setlength{\tabcolsep}{16.1pt}
  \centering
  {\scriptsize
  \begin{tabular}{lcccccc}
    \toprule
    & \multicolumn{2}{c}{Original Ensembled Model} & \multicolumn{4}{c}{Distilled Ensembled Model} \\ \cmidrule(lr){2-3} \cmidrule(lr){4-7}
    Inference Steps & 
    50 & 20 & 20 & 4 & 2 & 1 \\ \midrule
    IR & 
    0.4892 & 0.4803 & 0.4793 & \cellcolor{lred}{-0.5846} & \cellcolor{lred}{-0.5594} & \cellcolor{lred}{-0.5952} \\
    \bottomrule
  \end{tabular}}
  \caption{{Experimental results (IR) for distilling the ensembled model with AFA into fewer inference steps using LCM.}}
  \label{tab:reuslts-distill}
\end{table}

\section{{Evaluation on More Datasets}}

To validate the generality of our AFA, we evaluate it on additional datasets, which are DiffusionDB \citep{wang2022diffusiondb}, JourneyDB \citep{pan2023journeydb}, and LAION-COCO\footnote{\url{https://laion.ai/blog/laion-coco/}}, respectively.
We randomly selected 50,000 samples from each dataset to compare the performance of AFA against the base models and the baseline methods\footnote{Note that for JourneyDB, we ensured the selected samples are distinct from those in the training set.}.

The quantitative comparisons on the three additional datasets are presented in Table~\ref{tab:compare-on-diffusiondb}, Table~\ref{tab:compare-on-journeydb}, and Table~\ref{tab:compare-on-laion-coco}, respectively. 
The results demonstrate that our AFA consistently outperforms both the base models and the baseline methods, highlighting the generality and effectiveness of our approach.

\begin{table}[ht] \setlength{\tabcolsep}{3.2pt}
  \centering
  {\tiny
  \begin{tabular}{lcccccccccccccccc}
    \toprule
    & \multicolumn{8}{c}{Group I} & \multicolumn{8}{c}{Group II} \\ \cmidrule(lr){2-9} \cmidrule(lr){10-17}
    & FID $\downarrow$ & IS & CLIP-I & CLIP-T & AES & PS & HPSv2 & IR & FID $\downarrow$ & IS & CLIP-I & CLIP-T & AES & PS & HPSv2 & IR \\ \midrule
    Base Model A & 
    14.78 & 6.88 & .6433 & .2987 & 5.557 & 23.4113 & 29.9126 & .4354 &
    14.31 & 7.02 & .6578 & .3124 & 5.612 & 23.5527 & 30.1521 & .4413 \\
    Base Model B & 
    15.01 & 6.93 & .6513 & .3010 & 5.456 & 23.1123 & 29.8131 & .4223 &
    14.98 & 7.01 & .6618 & .3128 & 5.502 & 23.3128 & 30.0035 & .4307 \\
    Base Model C & 
    15.13 & 6.87 & .6441 & .2988 & 5.478 & 23.3423 & 29.9341 & .4339 &
    15.02 & 6.94 & .6391 & .2983 & 5.481 & 23.3420 & 29.9413 & .4291 \\ \midrule \midrule
    Wtd. Merging &
    17.41 & 6.37 & .6214 & .2776 & 5.213 & 23.2975 & 28.9938 & .4007 &
    16.98 & 6.81 & .6231 & .2984 & 5.501 & 23.2139 & 28.7310 & .4147 \\
    MBW &
    16.67 & 6.81 & .6378 & .2843 & 5.367 & 23.3002 & 28.8393 & .4115 &
    15.20 & 6.99 & .6392 & .2931 & 5.493 & 23.4412 & 29.0012 & .4216 \\
    autoMBW &
    16.66 & 6.76 & .6342 & .2663 & 5.377 & 23.3874 & 28.8432 & .4293 &
    15.17 & 7.00 & .6381 & .2891 & 5.551 & 23.3584 & 29.8413 & .4298 \\ \midrule
    MagicFusion & 
    14.77 & 6.98 & .6561 & .3123 & 5.674 & 23.4132 & 29.9241 & .4440 &
    14.21 & 7.02 & .6641 & .3125 & 5.611 & 23.5049 & 29.9941 & .4391 \\
    \textbf{AFA (Ours)} &
    \textbf{12.32} & \textbf{7.39} & \textbf{.6765} & \textbf{.3691} & \textbf{5.773} & \textbf{23.5132} & \textbf{30.0946} & \textbf{.4763} &
    \textbf{13.98} & \textbf{7.11} & \textbf{.6712} & \textbf{.3211} & \textbf{5.698} & \textbf{23.6130} & \textbf{30.1931} & \textbf{.4712} \\
    \bottomrule
  \end{tabular}}
  \caption{{Quantitative comparison in Group I and II on DiffusionDB.}}
  \label{tab:compare-on-diffusiondb}
\end{table}

\begin{table}[ht] \setlength{\tabcolsep}{3.2pt}
  \centering
  {\tiny
  \begin{tabular}{lcccccccccccccccc}
    \toprule
    & \multicolumn{8}{c}{Group I} & \multicolumn{8}{c}{Group II} \\ \cmidrule(lr){2-9} \cmidrule(lr){10-17}
    & FID $\downarrow$ & IS & CLIP-I & CLIP-T & AES & PS & HPSv2 & IR & FID $\downarrow$ & IS & CLIP-I & CLIP-T & AES & PS & HPSv2 & IR \\ \midrule
    \text{Base Model A} &
    17.80 & 5.63 & .5324 & .2484 & 4.520 & 19.2066 & 24.5997 & .3606 &
    17.21 & 5.75 & .5378 & .2513 & 4.577 & 19.1089 & 24.5196 & .3633 \\
    \text{Base Model B} &
    18.17 & 5.72 & .5379 & .2501 & 4.466 & 18.9965 & 24.5139 & .3458 &
    18.01 & 5.72 & .5428 & .2588 & 4.482 & 18.9069 & 24.3972 & .3483 \\
    \text{Base Model C} &
    17.98 & 5.62 & .5270 & .2434 & 4.507 & 19.1876 & 24.5743 & .3554 &
    17.87 & 5.61 & .5169 & .2398 & 4.445 & 18.9523 & 24.2945 & .3443 \\ \midrule \midrule
    \text{Wtd. Merging} &
    20.98 & 5.26 & .5170 & .2314 & 4.288 & 19.1237 & 23.8558 & .3327 &
    20.55 & 5.50 & .5108 & .2452 & 4.455 & 18.8782 & 23.3782 & .3399 \\
    \text{MBW} &
    19.74 & 5.58 & .5218 & .2365 & 4.399 & 19.1381 & 23.7068 & .3371 &
    18.35 & 5.74 & .5221 & .2378 & 4.484 & 19.0703 & 23.5806 & .3465 \\
    \text{autoMBW} &
    20.07 & 5.54 & .5204 & .2227 & 4.429 & 19.1960 & 23.6664 & .3508 &
    18.04 & 5.70 & .5182 & .2325 & 4.501 & 18.9612 & 24.2498 & .3510 \\ \midrule
    \text{MagicFusion} &
    17.76 & 5.76 & .5412 & .2591 & 4.672 & 19.2147 & 24.5909 & .3666 &
    16.93 & 5.72 & .5414 & .2573 & 4.550 & 19.0955 & 24.3816 & .3570 \\
    \textbf{AFA (Ours)} &
    \textbf{14.86} & \textbf{6.04} & \textbf{.5587} & \textbf{.3063} & \textbf{4.763} & \textbf{19.3324} & \textbf{24.7316} & \textbf{.3849} &
    \textbf{16.72} & \textbf{5.78} & \textbf{.5482} & \textbf{.2646} & \textbf{4.651} & \textbf{19.1986} & \textbf{24.5480} & \textbf{.3990} \\
    \bottomrule
  \end{tabular}}
  \caption{{Quantitative comparison in Group I and II on JourneyDB.}}
  \label{tab:compare-on-journeydb}
\end{table}

\begin{table}[ht] \setlength{\tabcolsep}{3.2pt}
  \centering
  {\tiny
  \begin{tabular}{lcccccccccccccccc}
    \toprule
    & \multicolumn{8}{c}{Group I} & \multicolumn{8}{c}{Group II} \\ \cmidrule(lr){2-9} \cmidrule(lr){10-17}
    & FID $\downarrow$ & IS & CLIP-I & CLIP-T & AES & PS & HPSv2 & IR & FID $\downarrow$ & IS & CLIP-I & CLIP-T & AES & PS & HPSv2 & IR \\ \midrule
    \text{Base Model A} &
    13.82 & 7.82 & .7328 & .3432 & 6.260 & 26.5012 & 33.8765 & .4962 &
    13.26 & 8.06 & .7549 & .3559 & 6.402 & 26.8726 & 34.4362 & .5080 \\
    \text{Base Model B} &
    14.12 & 7.89 & .7397 & .3467 & 6.164 & 26.1997 & 33.7818 & .4754 &
    13.87 & 8.05 & .7583 & .3617 & 6.286 & 26.5950 & 34.2635 & .4910 \\
    \text{Base Model C} &
    14.08 & 7.73 & .7280 & .3371 & 6.197 & 26.4257 & 33.8805 & .4886 & 
    13.84 & 7.90 & .7284 & .3379 & 6.254 & 26.6099 & 34.1549 & .4879 \\ \midrule \midrule
    \text{Wtd. Merging} &
    16.46 & 7.24 & .7101 & .3209 & 5.886 & 26.3613 & 32.8551 & .4573 &
    15.79 & 7.76 & .7149 & .3415 & 6.271 & 26.5130 & 32.8277 & .4762 \\
    \text{MBW} &
    15.45 & 7.68 & .7243 & .3275 & 6.071 & 26.3787 & 32.6951 & .4688 &
    14.11 & 8.07 & .7344 & .3327 & 6.311 & 26.7870 & 33.1185 & .4860 \\
    \text{autoMBW} &
    15.61 & 7.67 & .7218 & .3101 & 6.092 & 26.4808 & 32.6464 & .4852 &
    13.98 & 8.00 & .7328 & .3280 & 6.343 & 26.6503 & 34.0794 & .4946 \\ \midrule
    \text{MagicFusion} &
    13.80 & 7.92 & .7477 & .3594 & 6.425 & 26.5094 & 33.8799 & .5048 &
    13.13 & 8.05 & .7632 & .3632 & 6.421 & 26.8431 & 34.2797 & .5024 \\
    \textbf{AFA (Ours)} &
    \textbf{11.66} & \textbf{8.36} & \textbf{.7716} & \textbf{.4241} & \textbf{6.582} & \textbf{26.6673} & \textbf{34.1098} & \textbf{.5331} & 
    \textbf{12.84} & \textbf{8.16} & \textbf{.7717} & \textbf{.3735} & \textbf{6.552} & \textbf{26.9677} & \textbf{34.4844} & \textbf{.5405} \\
    \bottomrule
  \end{tabular}}
  \caption{{Quantitative comparison in Group I and II on LAION-COCO.}}
  \label{tab:compare-on-laion-coco}
\end{table}

\section{{Ensembling Models with Highly Correlated Features}}

As the number of the base models increases, reliance on multiple base models may leads to overfitting to correlated features.
To evaluate the robustness of our AFA against highly correlated features, we design an experiment.
Specifically, we selected one high-quality base model and one low-quality base model.
Using AFA, we ensemble the high-quality model with several low-quality models to evaluate whether the ensembled model will perform toward that of the low-quality model due to the dominance of the correlated features from the low-quality model.

As shown in Table~\ref{tab:compare-with-correlated-features}, although the performance the ensembled model slightly declines when ensembling with large number of low-quality models, it still outperforms the high-quality model.
This demonstrates that our AFA exhibits strong robustness to highly correlated features.

\begin{table}[ht] \setlength{\tabcolsep}{16.2pt}
  \centering
  {\scriptsize
  \begin{tabular}{lcccc}
    \toprule
    & AES & PS & HPSv2 & IR \\ \midrule
    High-Quality Base Model & 
    5.5641 & 21.8013 & 28.0183 & .4238 \\ 
    Low-Quality Base Model & 
    5.5013 & 21.4624 & 27.7246 & .2835 \\ \midrule 
    AFA (1 High-Quality Model + 1 Low-Quality Model) &
    5.6013 & 21.9341 & 28.3485 & .4746 \\
    AFA (1 High-Quality Model + 2 Low-Quality Models) &
    5.6000 & 21.9348 & 28.3398 & .4739 \\
    AFA (1 High-Quality Model + 3 Low-Quality Models) &
    5.5974 & 21.8248 & 28.3399 & .4730 \\
    AFA (1 High-Quality Model + 5 Low-Quality Models) &
    5.5831 & 21.8019 & 28.2035 & .4593 \\
    \bottomrule
  \end{tabular}}
  \caption{{Quantitative comparison of ensembling models with high-correlated features.}}
  \label{tab:compare-with-correlated-features}
\end{table}

\section{{Training with Dataset of Lower Quality}}

In this paper, we train our AFA using a high-quality dataset, JourneyDB. 
This raises the concern of whether AFA’s high performance is primarily attributed to the quality of the training dataset. 
To address this, we also train AFA using LAION-COCO, a dataset of lower quality compared to JourneyDB.

As shown in Table~\ref{tab:compare-training-laion-coco}, the model trained on LAION-COCO performs similarly to the one trained on JourneyDB, indicating that AFA is not sensitive to the quality of the training dataset. 
Its consistent performance across datasets of varying quality highlights its robustness and generalization ability. 
Regardless of whether the training dataset is high-quality or low-quality, AFA effectively leverages the features of the base models to deliver stable and reliable results.

\begin{table}[ht] \setlength{\tabcolsep}{2.95pt}
  \centering
  {\tiny
  \begin{tabular}{lcccccccccccccccc}
    \toprule
    & \multicolumn{8}{c}{Group I} & \multicolumn{8}{c}{Group II} \\ \cmidrule(lr){2-9} \cmidrule(lr){10-17}
    & FID $\downarrow$ & IS & CLIP-I & CLIP-T & AES & PS & HPSv2 & IR & FID $\downarrow$ & IS & CLIP-I & CLIP-T & AES & PS & HPSv2 & IR \\ \midrule
    \text{JourneyDB} &
    9.76  & 7.14 & .6926 & .2675 & 5.5201 & 21.8263 & 27.9734 & .4388 &
    10.27 & 7.42 & .6855 & .2717 & 5.5798 & 21.8059 & 28.0371 & .4892 \\
    \text{LAION-COCO} &
    9.74  & 7.16 & .6930 & .2664 & 5.5093 & 21.8257 & 27.9781 & .4390 &
    10.29 & 7.39 & .6861 & .2711 & 5.5789 & 21.8077 & 28.0363 & .4889 \\
    \bottomrule
  \end{tabular}}
  \caption{{Quantitative comparison in Group I and II trained on JourneyDB and LAION-COCO.}}
  \label{tab:compare-training-laion-coco}
\end{table}

\section{{More Experimental Details}}

We train our AFA in an environment equipped with 8 NVIDIA V100 GPUs, each with 32GB of memory.
When ensembling two base models, the GPU memory consumption is about 11GB under FP16 precision and a batch size of 1.
When ensembling three base models, the memory consumption increases to 14GB under the same settings.
For six base models, the memory consumption reaches about 30GB.
Due to the frozen parameters of the base models, training our AFA demands relatively modest GPU memory, even when ensembling a larger number of base models.

\end{document}